# Assessing the Physical Interpretability of AlphaEarth Foundation Embeddings Through Multi-Sensor Satellite Time-Series Benchmarking for Agricultural Mapping

El Houcine EL MOUSSAOUI[1,*], Chouaib EL HACHIMI[2,3,**], Saïd KHABBA[1,4], Salwa BELAQZIZ[4,5], Bouchra AIT HSSAINE[4], Abderrahman LAHROUNI[1], Salah ER-RAKI[4,6], Abdelghani CHEHBOUNI[4]

[1]LMFE, Department of Physics, Faculty of Sciences Semlalia, Cadi Ayyad University, Marrakesh, Morocco
[2]JC STEM Lab of Earth Observations, Department of Land Surveying and Geospatial Science, The Hong Kong Polytechnic University, Hung Hom, Kowloon, Hong Kong
[3]Research Centre for Artificial Intelligence in Geomatics, The Hong Kong Polytechnic University, Hung Hom, Kowloon, Hong Kong
[4]Center for Remote Sensing Applications, Mohammed VI Polytechnic University, Benguerir, Morocco
[5]LabSIV Laboratory, Faculty of Science, Department of Computer Science, Ibn Zohr University, Agadir, Morocco
[6]Laboratory of Excellence in Agrobiotechnology and Bioengineering, Center of Agrobiotechnology and Bioengineering, Faculty of Sciences and Techniques, Cadi Ayyad University, Marrakesh, Morocco
(*) email: elhou.moussaoui@gmail.com
(**) email: chouaib.elhachimi@polyu.edu.hk

## ABSTRACT

Accurate and dynamic maps of agricultural LULC are important for agricultural water resource management, particularly in areas where water scarcity is prevalent, such as Morocco. Here, we evaluate the utility of AlphaEarth Foundation Model (AEF) embeddings for agricultural LULC and irrigation-type mapping in Morocco. AEF embeddings were compared to single-date and time-series remote-sensing features using three machine-learning classifiers: XGBoost, CatBoost, and TabPFN. We also evaluated whether AEF embeddings can: (a) provide an efficient and transferable alternative to conventional direct use of Earth observation products and (b) reconstruct biophysical information. Results showed that multi-temporal feature time series achieved the highest performance, with accuracies of 0.93 for agricultural LULC and 0.87 for irrigation-type mapping when using TabPFN. However, AEF embeddings yielded comparable results, achieving accuracies of 0.91 for agricultural LULC and 0.85 for irrigation-type mapping, while reducing the need for extensive data processing and manual feature engineering typically required for satellite products. Spatial LULC mapping using TabPFN achieved comparable overall accuracies (0.85-0.86) across the three feature configurations, but revealed substantial differences in mapped tree cover, ranging from 14.57% for surface reflectance to 29.79% for AFE. Physical probing revealed that AEF embeddings strongly encode spectral indices (A30-SWIR: rho=0.58-0.67), (A38-NDMI: rho=-0.72), vegetation (A49-NDVI: rho = 0.72-0.73), and SAR structure (A25-VH: rho = 0.71). Future research should focus on enhancing the interpretability of embedding-based approaches by incorporating physical knowledge to develop more transparent and trustworthy AI systems for agricultural applications.

## 1 Introduction

Morocco's agricultural sector faces unexpected challenges, such as climate change, water scarcity, and the need for sustainable intensification. Since the 1960s, renewable water resources have declined by approximately 70%, and per-person water availability has fallen below the water scarcity threshold of 500 m³ per year (FAO, 2020). These challenges are exacerbated by unsustainable land use, population pressure, and limited irrigation infrastructure, with more than 60% of agricultural areas showing declining vegetation indices and rainfall variability exceeding a coefficient of variation of 100% (Alvarez and Govind, 2025). The Green Morocco Plan and its successors promoted water-efficient techniques such as drip irrigation, although this shift has, at times, inadvertently increased overall water intensity of production (Ouassanouan et al., 2023). Government targets aim to bring advanced systems such as sprinkler and drip irrigation to 55-60% of irrigated land (ROPUR, 2025). Morocco, a semi-arid country where agriculture accounts for more than 85% of national water consumption, is facing recurring droughts because of rising temperatures and changing precipitation patterns, which are impacting food security and living conditions in rural areas. Effective water resource management therefore requires the monitoring of different crop types and irrigation practices. However, the heterogeneity of cropping systems and irrigation methods complicates accurate and consistent monitoring across the country's agricultural landscapes (Dijk and Geurtsen, 2023). The crop basket consists of tree crops, cereals, and vegetables. In addition, farmers adopt multiple irrigation systems, including drip, sprinkler, and flood irrigation. In some regions, agriculture remains predominantly rainfed.

The increasing pressure on Morocco's water resources has intensified the need for accurate, spatially explicit information on crop water consumption and irrigation efficiency. Although ground-based surveys provide valuable point-based observations, they are not sufficient to capture the spatial and temporal variability of cropping systems across the country's heterogeneous agricultural zones. Satellite remote sensing overcomes this limitation by providing repeated, synoptic observations of agricultural landscapes. In particular, the complementary capabilities of Sentinel-1 SAR, Sentinel-2 multispectral imagery, and Landsat 8/9 optical data enable the characterization of soil moisture ,vegetation condition, and crop phenology, thereby providing essential information for informed water resource management. Agricultural landscapes are now covered almost sub-daily by open-access earth observation from multi-spectral and synthetic aperture radar (SAR) satellites. Together, the complementary views of canopy reflectance, vegetation structure, and soil moisture provided by Sentinel-1 (C-band SAR), Sentinel-2 (multispectral), and Landsat 8/9 constellations allow for the discriminative classification of crop varieties and irrigation techniques. Previous studies in Morocco have demonstrated the potential of remote sensing for crop-type mapping, irrigation monitoring, and agricultural water management in semi-arid and data-scarce environments (El-bouhali et al., 2023; El Moussaoui et al., 2025; Iounousse et al., 2015; Moumni and Lahrouni, 2021). Recent advances have extended beyond traditional optical imagery, with studies demonstrating the potential of C-band SAR data for monitoring soil moisture over wheat crops in semi-arid Morocco using Sentinel-1 backscattering coefficients and interferometric coherence (Ouaadi et al., 2020), and showing that synergetic use of Sentinel-1 and Sentinel-2 data enables soil moisture and irrigation mapping in semi-arid regions (Bousbih et al., 2018).

Irrigation practice identification from remote sensing is more challenging than crop-type mapping because different irrigation systems (drip, sprinkler, flood, rainfed, and gravity basin) can support similar crops and yield overlapping spectral and temporal signatures (Massari et al., 2021). The most discriminative features are soil moisture dynamics captured by Sentinel-1 SAR. Studies in the South Mediterranean region have shown that irrigation amounts and timing can be retrieved through data assimilation of surface soil moisture into the FAO-56 approach (Gao et al., 2017; Page et al., 2020). However, ambiguities between sprinkler and drip systems remain a persistent limitation across all sensors (Massari et al., 2021).

Recent research has explored the integration of optical and radar observations with auxiliary datasets, including thermal infrared data and derived statistical metrics, to improve the monitoring of vegetation health and soil moisture conditions (Sun and Kosmas, 2023). These advancements indicate a broader movement toward more comprehensive spatiotemporal characterization and enhanced data fusion in both land use/land cover (LULC) and irrigation classification. However, the majority of these approaches continue to rely on manually constructed features, such as spectral indices and radar backscatter coefficients, which are inherently tied to specific sensors, calibrated for particular regions, and often lack transferability across different environmental contexts. Furthermore, inconsistencies in spatial sampling, observation frequency, and physical measurement principles across different satellite platforms present significant obstacles to effective data integration and temporal alignment.

Despite the wealth of Earth observation data now available, agricultural monitoring still faces major obstacles, including data heterogeneity, limited availability of labeled training data, and the poor generalizability of models trained on specific regions or time periods. Conventional features like whether derived from optical reflectance or radar backscatter, are typically designed for individual sensors and do not adequately encode the temporal information critical for capturing crop phenology and irrigation patterns. The growing complexity of these systems has driven interest in geospatial foundation models that can autonomously extract meaningful representations from massive, unlabeled satellite data archives.

Recent developments in geospatial foundation modeling present a promising avenue to overcome the limitations of traditional feature engineering in remote sensing. AlphaEarth Foundation (AEF) offers a transformative approach, leveraging a 480-million-parameter model pre-trained on approximately 3 billion images from 5 million locations to learn a unified representation of Earth's surface that synthesizes spatial, spectral, and temporal information across multiple sensor sources (DeepMind, 2025; Jakubik et al., 2026) . During inference, AEF ingests annual multi-source observations including Sentinel-2 and Landsat optical data, Sentinel-1 SAR backscatter, LiDAR, and climate simulations, and compresses them into 64-dimensional embedding vectors at 10 m resolution (Brown et al., 2025). These embeddings have shown consistent superiority over handcrafted features across diverse geospatial tasks, reducing average prediction errors by 24% (Vinge et al., 2025). Furthermore, AEF embeddings are readily available as analysis-ready datasets on Google Earth Engine, allowing researchers to bypass the substantial preprocessing and computational infrastructure typically required for foundation model deployment (Feng et al., 2026). This combination of strong performance, compact representation, and ease of access makes AEF particularly attractive for agricultural monitoring applications in Morocco and other semi-arid regions, where the ability to derive robust, transferable features without extensive feature engineering is essential for operational-scale LULC and irrigation mapping.

Google Earth Engine (GEE) enables the efficient processing of large spatiotemporal datasets without requiring local data storage or high-performance computing infrastructure(Amani et al., 2020). This has enabled the construction of complex feature engineering workflows within a single computational environment, without requiring local data downloads or storage. Nevertheless, despite this technical infrastructure, two systematic methodological gaps persist in the remote sensing crop mapping literature. First, most published works rely on fixed seasonal or monthly composites that aggregate imagery without regard to the specific observation date of ground truth samples. When field training data are collected at heterogeneous dates spanning multiple months, this temporal misalignment between label and feature can introduce substantial phenological noise, degrading classification accuracy, particularly for crops with rapid developmental cycles such as wheat and barley (Griffiths et al., 2019). Second, time-series frameworks designed for phenological monitoring often process each sensor independently, losing the multi-source complementarity that distinguishes between crops with similar optical signatures but different structural properties, a distinction that SAR backscatter can resolve (Veloso et al., 2017).

The choice of machine learning classifier is a critical determinant of mapping accuracy in LULC and irrigation classification. Among the available algorithms, gradient boosting machines have emerged as the leading approach, offering robust performance, interpretability, and efficient handling of mixed data types. XGBoost has been widely adopted for land cover classification, leveraging regularized gradient boosting to achieve high accuracy across diverse remote sensing datasets (Chen and Guestrin, 2016).

CatBoost enhances this framework through ordered boosting, which reduces overfitting and improves generalization on categorical features, making it particularly suitable for agricultural applications (Prokhorenkova et al., 2018). A more recent alternative, TabPFN, introduces a transformer-based architecture pre-trained on synthetic data that performs Bayesian in-context learning without gradient-based optimization, offering particular advantages for small and imbalanced datasets (Hollmann et al., 2023b). While these classifiers have been applied individually to agricultural mapping, comparative evaluations across multiple feature modalities, from handcrafted spectral indices to AlphaEarth embeddings, remain limited.

Building on the identified gaps, this study pursues three interrelated objectives. First, we compare the performance of AlphaEarth neural embeddings against single-date and time-series satellite features for LULC and irrigation classification in Moroccan irrigated perimeters. We evaluate the relative effectiveness of three machine learning classifiers (XGBoost, CatBoost, and TabPFN) across five feature modalities: AlphaEarth embeddings, single-date surface reflectance, single-date spectral indices, multi-temporal surface reflectance time series, and multi-temporal spectral index time series. Second, we assess whether AlphaEarth embeddings can serve as a viable, efficient alternative to traditional handcrafted features, potentially reducing the need for extensive feature engineering while maintaining or improving classification accuracy. Third, we perform an explainability analysis by combining SHAP attribution with physical probing of EO descriptors from the 64-dimensional latent space via Spearman correlation, Ridge, and Random Forest probes to identify predictive drivers and quantify the recoverability of biophysical information. All models are evaluated using leave-one-out validation (LOOV) and stratified 5-fold cross-validation (5CV) to ensure robust and comparable performance estimates. Through this comprehensive benchmarking, we aim to provide evidence-based guidance for the design of operational LULC and irrigation monitoring systems in semi-arid agricultural environments.

## 2 Study area and ground truth data

The study covers most agricultural regions in Morocco (Figure 1), namely Tangier-Tétouan-Al Hoceïma, Rabat-Salé-Kénitra, Béni Mellal-Khénifra, Casablanca-Settat, Marrakech-Safi, and Souss-Massa. These six regions were surveyed between 2023 and 2025 by a team of professors, PhD students, and technicians from the Center for Remote Sensing Applications at Mohammed VI Polytechnic University (Ouassanouan et al., 2026). These surveys collected georeferenced field boundaries along with detailed attributes describing crop type, crop status, irrigation method in use at the time of acquisition, and other relevant agronomic characteristics in addition to broader agricultural management practices.

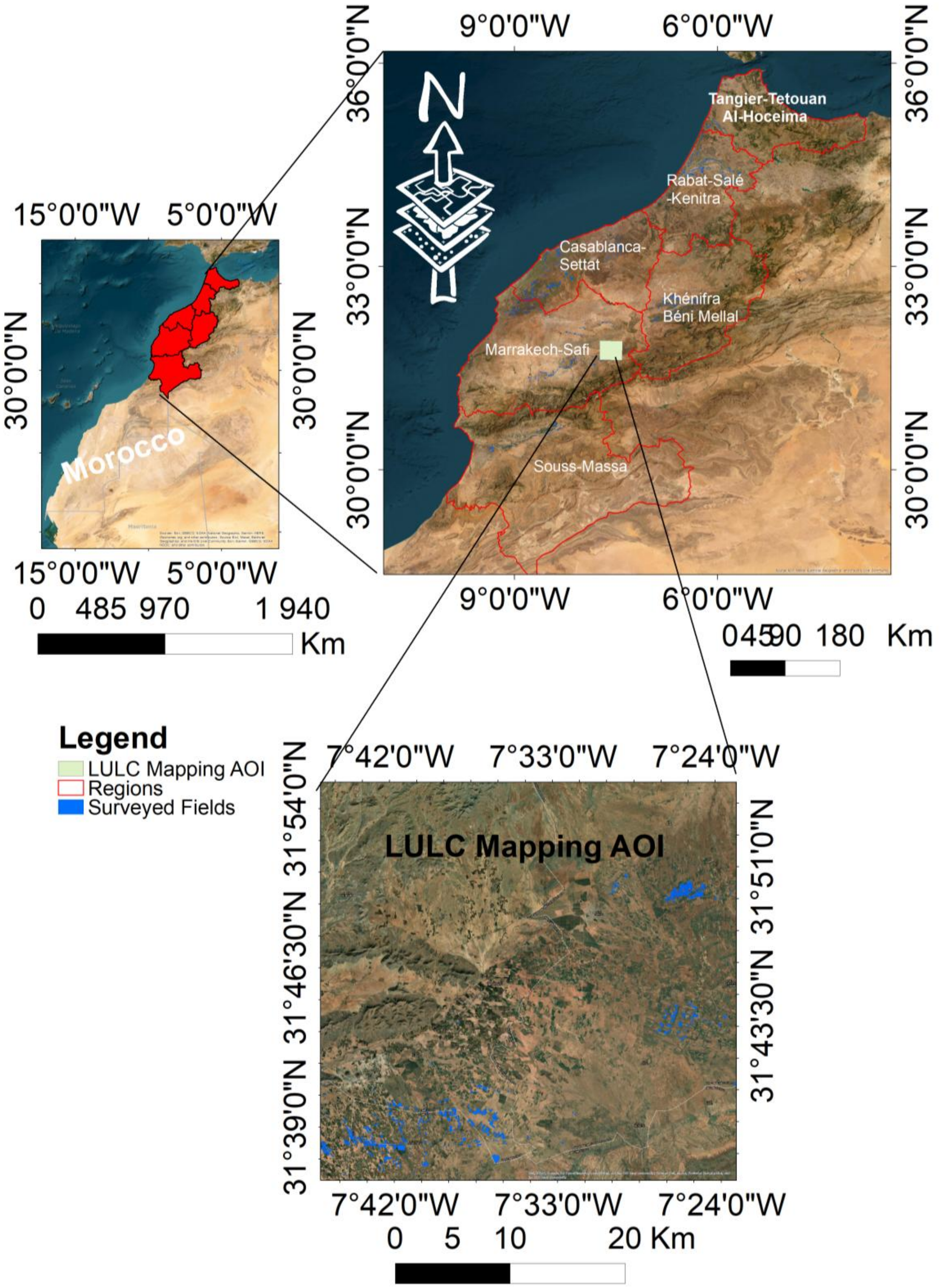

*Figure 1: Spatial distribution of surveyed agricultural fields across Morocco (2023-2025).*

Ground reference data for model training and validation were collected through field surveys across six agricultural zones in Morocco: Doukkala (4,405 polygons), Gharb (1,866), El Haouz (1,752), Tadla (1,558), Souss (517), and Statte (285), totaling 10,383 georeferenced polygons covering approximately 18,167 ha. Field campaigns were conducted between December 2023 and January 2025, with the largest collection efforts occurring in April 2024 (n = 3,063), December 2024 (n = 2,200), and August 2024 (n

= 1,866), reflecting sampling across multiple phenological stages and growing seasons. Each parcel was labeled along two independent axes: crop type/land cover and irrigation practice. Following quality control, labels were standardized into four LULC classes (bare soil, seasonal crop, trees, and trees + seasonal crop) and four irrigation classes ( drip irrigation, flood irrigation, sprinkler irrigation, and rainfed). A small number of classes were excluded from the curated dataset either due to trivial spectral separability (water, weed) or insufficient sample size (pivot irrigation, basin irrigation, and a duplicate-label data entry artifact, flood irrigation) (Table 1).

*Table 1. Distribution of LULC classes and irrigation types in the ground truth dataset.*

| Category | Class | Count | Percentage |
|---|---|---|---|
| **LULC Type** | Seasonal Crop | 6,975 | 67.19% |
| | Trees | 1,427 | 13.74% |
| | Bare soil | 942 | 9.07% |
| | Trees + Seasonal | 385 | 3.71% |
| **Irrigation Type** | Flood Irrigation | 3,022 | 29.11% |
| | Drip Irrigation | 2,890 | 27.83% |
| | Sprinkler Irrigation | 1,245 | 11.99% |
| | Rainfed | 1,216 | 11.71% |

## 3 Remote sensing data and preprocessing

This study integrated three satellite datasets, reflecting the growing use of multi-sensor approaches in agricultural monitoring. These sources provide spectral, structural, temporal, and learned landscape representations that together support characterizing crop types and irrigation systems in Moroccan agricultural areas.

### 3.1 Sentinel-2 images and pre-processing

The S2 satellite system consists of two satellites: Sentinel-2A, launched on 23 June 2015, and Sentinel-2B, launched on 7 March 2017. The system provides high-resolution imagery, with most bands at 10 or 20 meters. After 7 March 2017, the revisit time at the equator is five days, and in some mid-latitude regions, it is reduced to two to three days. Sentinel-2 imagery was obtained from the Copernicus collection, which supplies Level-2A bottom-of-atmosphere (BOA) surface reflectance data. These data are processed using the Sen2Cor atmospheric correction processor and cross-sensor harmonization to ensure radiometric consistency between Sentinel-2A and Sentinel-2B. S2 preprocessing follows the procedures described by (García-Ontiyuelo et al., 2026). For each field parcel, imagery was spatially filtered to the parcel boundary and temporally constrained to a ±15-day window centered on the parcel's recorded survey date. A scene-level cloud filter was applied, retaining only acquisitions with a cloudy pixel percentage below 30% to minimize cloud contamination. All bands at 10-meter native spatial resolution were rescaled by a factor of 0.0001 to convert integer digital numbers to physical surface reflectance values in the range [0, 1].

### 3.2 Landsat 8 and 9 Operational Land Imager

Landsat 8 (launched in 2013) and Landsat 9 (launched in 2021) imagery was sourced from the USGS Collection 2 Level-2 surface reflectance product. Atmospheric correction was conducted using the Land Surface Reflectance Code (LaSRC) algorithm (Vermote et al., 2016), , which ensured radiometrically consistent bottom-of-atmosphere (BOA) reflectance values for both sensors. Spatial and temporal filtering consistent with that applied to Sentinel-2 was implemented for the Landsat data: acquisitions were limited to a ±15-day window around the survey date and filtered for cloud cover. Per-pixel median compositing was performed on all qualifying scenes. Six surface reflectance bands were retained: Blue, Green, Red, Near-Infrared (NIR), Shortwave Infrared 1 (SWIR1), and Shortwave Infrared 2 (SWIR2), each at a native spatial resolution of 30 meters.

## 3.3 Sentinel-1

Sentinel-1 (S1A and S1B) carries a C-band SAR instrument operating at 5.405 GHz (wavelength approximately 5.6 cm), providing all-weather, day-and-night imaging capability unaffected by cloud cover a critical advantage during Morocco's rainy growing season (November-April) when optical data are frequently unavailable. C-band radar is sensitive to canopy volume scattering at intermediate vegetation densities, making it particularly responsive to crop structure changes during emergence, vegetative growth, heading, and senescence.

Sentinel-1 GRD data in GEE are provided in gamma-naught backscatter in decibels (dB). The preprocessing pipeline first converts backscatter from dB to linear power units using the inverse logarithmic transformation: $\sigma_linear = 10^{(\sigma^0(dB)\,/\,10)}$, applied independently to VV and VH polarization channels. Speckle reduction is a critical preprocessing step for SAR data analysis, given the coherent noise that characterizes GRD products. To effectively reduce speckle while preserving edges and fine structures, a 3×3 Sigma Lee filter is applied in linear units (El Moussaoui et al., 2024). This filter is chosen for its adaptive properties, which outperform standard mean filtering by avoiding over-smoothing. Processing in linear power space preserves radiometric unbiasedness, as averaging in linear units is statistically correct whereas averaging in dB introduces a multiplicative bias. After filtering, backscatter is reconverted to dB for index computation and export, yielding speckle-reduced VV and VH backscatter in decibels.

*Table 2. Summary of radiometric preprocessing transformations applied per sensor*

| Sensor | Input format | Transformation | Output unit |
|---|---|---|---|
| **Sentinel-2** | DN (integer, ×10,000) | Not Needed | Surface reflectance [0-1] |
| **Landsat 8/9** | DN (integer) | Not Needed | Surface reflectance [0-1] |
| **Sentinel-1** | σ° (dB, GRD) | Linear to mean filter to dB | Speckle-filtered σ° (dB) |

## 3.4 Optical spectral indices

Following radiometric preprocessing, a unified feature engineering procedure was applied independently to Sentinel-2 and Landsat 8/9 observations using six corresponding spectral bands (Blue, Green, Red, NIR, SWIR1, and SWIR2). Despite differences in their native spatial resolutions (10-20 m for Sentinel-2 and 30 m for Landsat 8/9), spectral indices were calculated separately for each sensor to retain their respective spectral characteristics. Nine vegetation and land-surface indices were selected to characterize canopy greenness and vigor, vegetation water content, soil background effects, bare soil exposure, and surface water presence (Table 3). Two additional brightness-related features were computed: VIS-sum (Blue + Green + Red) and Total-sum (VIS-sum + NIR). Monthly median composites were subsequently generated independently for each sensor, and the resulting features were integrated into a unified dataset for agricultural LULC and irrigation-type classification.

*Table 3. Spectral indices computed from Sentinel-2 and Landsat 8/9. All indices are computed identically for both sensors.*

| Index | Formula | Crop mapping relevance and irrigation type identification |
|---|---|---|
| Normalized Difference Vegetation Index (NDVI) | $(NIR - RED)/(NIR + RED)$ | Primary greenness indicator, tracks crop emergence to senescence (Tucker, 1979). |
| Normalized Difference Moisture Index (NDMI) | $(NIR - SWIR1)/(NIR + SWIR1)$ | Drought stress, irrigation status (Jin and Sader, 2005). |
| Normalized Difference Water Index (NDWI) | $(GREEN - NIR)/(GREEN + NIR)$ | Irrigated field boundary delineation, important for maintaining the contrast between |
| | | water and vegetation (Gao, 1996). |
| Green Normalized Difference Vegetation Index (GNDVI) | $(NIR - GREEN)/(NIR + GREEN)$ | Saturates less than NDVI at high LAI (Gitelson and Mark N, 1998). |
| Enhanced Vegetation Index (EVI) | $2.5 \times (NIR - RED)/(NIR + 6RED - 7.5BLUE + 1)$ | Handles NDVI saturation in dense canopy and aerosol-resistant (Huete et al., 2002). |
| Modified Soil Adjusted Vegetation Index (MSAVI) | $(2NIR + 1 - \sqrt{[(2NIR + 1)^2 - 8(NIR - RED)]})/2$ | Early growth stage detection and effectively isolates vegetation signal from bare soil (Qi et al., 1994). |
| Soil Adjusted Vegetation Index (SAVI) | $((NIR - RED)/(NIR + RED + 0.5)) \times 1.5$ | Semi-arid croplands and high soil background contribution (Huete, 1988) |
| Visible Atmospherically Resistant Index (VARI) | $(GREEN - RED)/(GREEN + RED - BLUE)$ | Atmospheric-resistant and field-level greenness fraction (Gitelson et al., 2002) |
| Bare Soil Index (BSI) | $((SWIR1 + RED) - (NIR + BLUE))/((SWIR1 + RED) + (NIR + BLUE))$ | Soil characterization and conservation agriculture monitoring (Fernández-Buces et al., 2006) |

## 3.5 SAR derived features

Four radar features are derived from each Sentinel-1 acquisition following speckle filtering (Table 4). The VH/VV polarization ratio is a dimensionless index sensitive to volumetric scattering relative to surface scattering, increasing with vegetation canopy density and complexity. The Radar Vegetation Index (RVI) provides a normalized measure of canopy randomness and has been validated as a robust indicator of crop biomass accumulation and phenological stage in C-band SAR data (Arai et al., 2025). RVI values approach 0 for smooth bare surfaces (dominated by VV specular reflection) and approach 1 for complex, fully developed canopies with dominant volume scattering.

The complementarity between optical and SAR features is particularly valuable for the Moroccan agricultural context: cereal crops in their vegetative phase can be difficult to distinguish from green pasture using optical indices alone, but their different canopy architectures produce distinct SAR backscatter signatures, especially in VH polarization.

*Table 4. Radar-derived features computed from Sentinel-1 GRD data.*

| Feature | Formula / Processing | Physical interpretation |
|---|---|---|
| VV (filtered) | Mean filter (3×3) in linear, reconverted to dB | Surface roughness, soil moisture, and specular reflection from crop rows. |
| VH (filtered) | Mean filter (3×3) in linear, reconverted to dB | Volume scattering, canopy structure, and sensitive to vegetation density. |
| VH/VV ratio | $\sigma(VH) / \sigma(VV)$ | Depolarization ratio; increases with canopy complexity and randomness. |
| RVI | $4 \times \sigma(VH) / (\sigma(VH) + \sigma(VH))$ | Normalized radar vegetation index: 0 = bare soil, 1 = dense canopy. |

## 3.6 AlphaEarth embeddings data

In addition to the engineered multi-sensor features derived from Sentinel-2, Landsat 8/9, and Sentinel-1, this study incorporates pre-computed embeddings from the AlphaEarth foundation model. AlphaEarth is a large-scale Earth Observation foundation model trained on a globally diverse, multi-sensor, multi-temporal dataset spanning optical, SAR, and ancillary data streams. The 64-dimensional AlphaEarth satellite embeddings (bands A00-A63) are retrieved for each field polygon through GEE. These embeddings encode information from a wide range of EO sources, including optical data (Sentinel-2 and Landsat-8), thermal data (Landsat-9), microwave data (C-band SAR from Sentinel-1 and L-band SAR from ALOS, PALSAR, and ScanSAR), elevation data (Copernicus DEM), LiDAR (GEDI), climate data (ERA5-Land), gravity field data (GRACE), land cover data, and even textual data from Wikipedia and GBIF. Each embedding vector was matched with the corresponding field attributes, resulting in two tabular datasets: (i) the irrigation-type retrieval dataset and (ii) the crop-type classification dataset. These datasets served as input to the predictive models evaluated in this study (Brown et al., 2025).

The AlphaEarth embeddings are accessed as pre-exported GeoTIFF tiles aligned to the GEE projection grid, and polygon-level values are extracted using a spatial join against the centroid coordinates of each deduplicated ground truth sample. No additional temporal compositing, radiometric normalization, or feature scaling is applied to the embedding vectors before classification, as the foundation model's internal normalization is assumed to produce embeddings on a consistent latent scale across all samples (EL Hachimi et al., 2026).

# 4 Methods

## 4.1 Tabular learners

A central design decision of this framework is to benchmark classification performance across three fundamentally distinct tabular learning paradigms that represent the current state of the art for structured, feature-table-based prediction tasks. This choice is motivated by the observation that no single algorithm consistently dominates across varying sample sizes, feature dimensionalities, and class distributions in remote sensing classification literature. The three classifiers selected (XGBoost, CatBoost, and TabPFN) differ not only in their training algorithms but in their underlying statistical frameworks, making their comparison informative about the nature of the classification problem itself.

XGBoost is a scalable gradient-boosted decision tree ensemble that constructs an additive sequence of shallow regression trees, each correcting the residuals of the preceding ensemble (Chen and Guestrin, 2016). It optimizes a regularized objective function using second-order Taylor approximations of the loss, enabling efficient handling of large, high-dimensional feature tables with sparse structures. In this study, XGBoost was configured to run on CPU with multi-threaded parallelism, with the number of workers determined dynamically at runtime based on available hardware. Its widespread use as a baseline in remote sensing classification makes it a natural reference point for evaluating the other classifiers.

CatBoost is a gradient boosting variant specifically designed to handle categorical features without preprocessing through an ordered boosting scheme that prevents target leakage during training (Prokhorenkova et al., 2018). Although the satellite-derived features in this study are entirely numerical, CatBoost's ordered boosting strategy also provides variance reduction benefits for numerical targets by computing leaf statistics on a permuted subset of the training data at each boosting iteration.

TabPFN (Hollmann et al., 2023a) represents a fundamentally different learning paradigm: it is a prior-fitted transformer network trained offline on millions of synthetically generated small tabular classification problems drawn from a prior distribution over Bayesian networks. At inference time, TabPFN treats the entire training set as an in-context prompt and performs a single forward pass through the transformer to produce class probability estimates for all test samples simultaneously, without any gradient-based optimization on the target dataset. This amortized Bayesian inference mechanism enables TabPFN to approximate the posterior predictive distribution over class labels conditioned on the observed training set, making it particularly well suited to datasets with moderate sample sizes (hundreds to low thousands) and a high feature-to-sample ratio, as is typical for field-survey-derived remote sensing training sets.

## 4.2 Dual-mode temporal compositing

1) In the Point-in-Time mode, the temporal extent of image retrieval is individually defined for each ground truth polygon based on its associated survey date. A symmetric window of +/-15 days is constructed around the survey date, and all three sensor collections are filtered to this window independently. The 15-day half-window balances image availability, particularly for Sentinel-2, which may have fewer cloud-free acquisitions in winter months, against phenological representativeness.

2) In the Monthly Time Series mode, a fixed sequence of 36 monthly compositing windows covering January 2023 to December 2025 is generated for all three sensor collections simultaneously. This mode captures the full seasonal phenological trajectory of each sample, enabling the construction of time-series feature stacks with up to 15 × number of dates over 36 months per sample for the surface reflectance configuration, and up to 19 × number of dates over 36 months for the spectral indices configuration.

## 4.3 Dataset configurations and benchmark design

A systematic benchmark was constructed across ten dataset configurations corresponding to five distinct feature representation types (Table 5), each evaluated for crop type classification (LULC target: bare soil, trees + seasonal, trees, seasonal crop) and irrigation system identification (Irrigation target) where applicable. The five feature types are: (i) AlphaEarth foundation model (Embeddings) (64-dimensional, annual); (ii) single-date surface reflectance (15 bands matched to survey date via point-in-time mode); (iii) single-date spectral indices (19 indices, same temporal alignment); (iv) time-series surface reflectance (15 × number of dates); and (v) time-series spectral indices (19 × number of dates) over the 36-month study period.

*Table 5. Dataset configurations evaluated in the machine learning benchmark.*

| Dataset type | Target | Features | Sensors | Temporal scope |
|---|---|---|---|---|
| AlphaEarth Embedding | LULC + Irrigation | 64 dimensions | Multi-sensor | Full |
| Single-date (surface reflectance) | LULC + Irrigation | 15 bands | S2 + L89 + S1 | One date per sample |
| Single-date (spectral indices) | LULC + Irrigation | 19 indices | S2 + L89 + S1 | One date per sample |
| Time-series (surface reflectance) | LULC + Irrigation | 15 x dates | S2 + L89 + S1 | 36 months |
| Time-series (spectral indices) | LULC + Irrigation | 19 x dates | S2 + L89 + S1 | 36 months |

## 4.4 Evaluation Protocol

Two complementary evaluation protocols were applied to each of the thirty classifier-dataset-target combinations :

- ✓ **Leave-one-out validation (LOOV):** each model is trained on the full dataset and evaluated via leave-one-out cross-validation, producing per-class precision, recall, and F1-score along with overall accuracy. This protocol reflects the standard single-split performance characteristic.
- ✓ **5-Fold cross-validation (CV5):** the dataset is partitioned into five stratified folds. Training and evaluation are repeated five times with each fold held out once. Mean accuracy (Eq. (1)), precision (Eq. (2)), recall (Eq. (3)), and F1-score (Eq. (4)) are reported across folds, providing a statistically robust performance estimate with reduced variance due to data splitting.

$$Accuracy = \frac{\sum_{i=1}^{N} X_{ii}}{\sum_{i=1}^{N} \sum_{j=1}^{N} X_{ij}} \tag{1}$$

where N is the number of rows in the error matrix, $X_{ii}$ is the number of observations in the diagonal elements (true positives for each class). $X_{ij}$ the number of observations in row i and column j (the total number of observations is the sum of all TP, FP, and FN across all classes).

$$Precision_c = \frac{TP_c}{(TP_c + FP_c)} \tag{2}$$

$$Recall_c = \frac{TP_c}{(TP_c + FN_c)} \tag{3}$$

where TPc (True Positives for class c) counts correctly predicted instances of class c, FPc (False Positives for class c) counts instances incorrectly predicted as class c, and FNc (False Negatives for class c) counts instances of class c that were missed (incorrectly predicted as other classes).

$$F1 - score = 2 \times \frac{Precision_c \times Recall_c}{(PPrecision_c + Recall_c)} \tag{4}$$

Label encoding mapped string class labels to contiguous integer indices (0 to K-1) using pandas factorize. Rows with invalid or unmapped labels were excluded.

## 4.5 Explainability analysis of EO representations

To investigate the contribution of different Earth-observation representations to agricultural LULC and irrigation-type classification, an explainability analysis was performed using two complementary feature attribution approaches: SHAP (SHapley Additive exPlanations) (Lundberg and Lee, 2017) and intrinsic XGBoost feature importance. The objective was not only to identify the most influential predictive variables but also to compare the interpretability characteristics of latent foundation-model embeddings against conventional physically interpretable EO variables.
For each classification task, three feature representations were independently evaluated:

1. AEF embeddings, consisting of 64-dimensional learned latent representations $z = [A00, A01, \ldots, A63]$.
2. Raw Optical + SAR representation, including Landsat 8/9 and Sentinel-2 surface reflectance bands together with Sentinel-1 SAR backscatter and derived radar variables.
3. Spectral-index representation, including vegetation, moisture, and soil-sensitive indices derived from optical observations.

A gradient-boosted decision tree classifier based on XGBoost was trained separately for each representation and target task. The trained models were subsequently analyzed using global feature attribution. For a prediction model $f(x)$, SHAP decomposes the model output as using equation (5).

$$f(x) = \phi_0 + \sum_{j=1}^{M} \phi_j \tag{5}$$

where $\phi_j$ represents the contribution of feature $j$ to the prediction relative to the expected model output $\phi_0$. Global feature importance was calculated using the mean absolute SHAP value (Eq. (6)).

$$I_j^{SHAP} = \frac{1}{N} \sum_{i=1}^{N} |\phi_{i,j}| \tag{6}$$

where N represents the number of samples and j represents the input feature.

In addition to SHAP attribution, the native XGBoost feature importance was calculated based on the accumulated contribution of each variable to tree construction. This provides a model-internal measure of feature utilization based on split decisions and gain.

The two approaches were analyzed independently because they provide complementary perspectives: SHAP quantifies the influence of each feature on final model predictions, based on a game-theory approach, whereas XGBoost importance, in turn, describes how frequently and effectively variables contribute to decision-tree optimization.
For each task and representation, the top-ranked features were extracted and visualized separately for SHAP and XGBoost importance to facilitate comparison between latent and physically interpretable feature spaces.

## 4.6 Physical probing of AlphaEarth embeddings

To investigate whether the latent AEF representation retains physically interpretable Earth-observation information, the 64 latent embeddings were used to reconstruct conventional optical, spectral-index, and SAR variables. The original 1,224 multi-temporal EO features were summarized using eight temporal descriptors, namely mean, standard deviation, minimum, maximum, amplitude, 10th percentile, 90th percentile, and timing of the maximum response, yielding 272 physical probing targets that represent different environmental properties, including vegetation condition, moisture availability, spectral response, soil characteristics, SAR structural information, and temporal behavior. Two complementary probe models were then fitted: Ridge regression as a linear probe to assess directly accessible information in the latent space, and Random Forest regression as a nonlinear probe to determine whether additional information could be recovered through nonlinear relationships among embedding dimensions.
This physical probing experiment aims to move beyond predictive accuracy of AEF and ask a complementary question: to what extent can conventional, physically interpretable Earth-observation information be recovered from the AEF latent space? Conceptually, the analysis treats the embedding as a compressed representation ($z \in R^{64}$), (Eq. (7)), and evaluates whether observable EO variables ($y^{(p)}$), such as NDVI, NDMI, SWIR reflectance, or Sentinel-1 backscatter, can be reconstructed from ($z$) (Eq. (8)).

$$Z = [A_{00}, A_{01}, \dots, A_{63}], \tag{7}$$

$$R\widehat{y^{(p)}} = f_p(z), \tag{8}$$

where ($f_p(\cdot)$) denotes a probe trained specifically for physical variable (p). A high out-of-sample reconstruction performance indicates that information related to that physical quantity remains accessible within the embedding, without implying that any individual AEF dimension has a unique one-to-one biophysical interpretation.
The probing targets were derived from the conventional multi-sensor EO representation used in this study. They comprised Landsat 8/9 and Sentinel-2 surface-reflectance bands, vegetation and moisture-sensitive spectral indices, soil/background indicators, and Sentinel-1 SAR variables describing backscatter and vegetation/structural response. Because each EO variable was represented by 36 temporal observations, directly probing every individual date would produce a highly redundant target space and would emphasize acquisition-specific variation rather than persistent environmental information. We therefore summarized each temporal sequence (Eq. (9)) using eight temporal descriptors designed to capture both the general surface state and its temporal variability. These included the temporal mean (Eq. (10)), the standard deviation (Eq. (11)), minimum and maximum (Eq.

(12)), temporal amplitude (Eq. (13)), the 10th and 90th percentiles (Eq. (14)), and the timing of the maximum observed response (Eq. (15)).

$$x^{(p)} \left[x_1^{(p)}, x_2^{(p)}, \ldots, x_T^{(p)}\right], \qquad T = 36 \tag{9}$$

$$\mu_p = \frac{1}{T_p} \sum_{t \in \mathcal{V}_p} x_t^{(p)}, \tag{10}$$

$$\sigma_p = \sqrt{\frac{1}{T_p - 1} \sum_{t \in \mathcal{V}_p} \left(x_t^{(p)} - \mu_p\right)^2}, \tag{11}$$

$$x_{p,\min} = \min_t x_t^{(p)}, \qquad x_{p,\max} = \max_t x_t^{(p)}, \tag{12}$$

$$A_p = x_{p,\max} - x_{p,\min}, \tag{13}$$

$$Q_{0.10}^{(p)} \quad \text{and} \quad Q_{0.90}^{(p)}, \tag{14}$$

$$\left(x_{\tau_p}^{(p)} = \max_t x_t^{(p)}\right) \tag{15}$$

$\mathcal{V}_p$ denotes the set of valid, non-missing observations for variable ($p$), allowing the temporal summaries to remain robust to incomplete observations. The transformation reduced the original 1,224 multi-temporal EO variables to 272 compact physical probing targets. Agricultural LULC and irrigation-type labels were deliberately excluded because the purpose of this analysis was to probe recoverable EO information rather than downstream class information.

We first examined the organization of the latent space using Spearman rank correlation. For each AlphaEarth dimension ($A_j$) and aggregated physical variable ($y^{(p)}$), (Eq. (16)).

$$\rho_{j,p} = \rho_s\left(A_j, y^{(p)}\right) \tag{16}$$

This analysis identifies embedding dimensions whose variation is monotonically associated with conventional EO observables. Because a single latent dimension may encode multiple correlated environmental properties, these relationships were interpreted as statistical associations rather than direct physical semantics.

Correlation analysis alone, however, cannot determine whether a physical quantity is distributed across multiple embedding dimensions. We therefore complemented the dimension-wise analysis with predictive probing using the full 64-dimensional representation. Two probe families were considered. First, Ridge regression was used as a linear probe (Eq. (17)), with coefficients estimated by minimizing (Eq. (18)).

$$\widehat{y^{(p)}} \beta_0 + \sum_{j=1}^{64} \beta_j A_j, \tag{17}$$

$$\mathcal{L}_{Ridge} = \sum_{i=1}^{N} \left(y_i^{(p)} - \widehat{y_i^{(p)}}\right)^2 + \lambda \sum_{j=1}^{64} \beta_j^2 \tag{18}$$

The Ridge probe was intentionally included because successful linear recovery provides stronger evidence that the physical information is readily accessible in the latent representation rather than requiring complex nonlinear decoding. Second, a Random Forest regressor was used as a nonlinear probe to determine whether additional information could be recovered through interactions or nonlinear combinations of AEF dimensions. The comparison between the two probes therefore provided insight not only into whether a physical variable was encoded, but also into how accessible that information was within the latent space.

Recoverability of physical variables from AEF embeddings is quantified using independent holdout samples and regression metrics, including $R^2$ (Eq. (19)), RMSE (Eq. (20)), and MAE (Eq. (21)).

$$R^2 = 1 - \frac{\sum_{i=1}^{n}(y_i - \hat{y}i)^2}{\sum i = 1^n (y_i - \bar{y})^2} \tag{19}$$

$$RMSE = \sqrt{\frac{1}{n}\sum_{i=1}^{n}(y_i - \hat{y}_i)^2} \tag{20}$$

$$MAE = \frac{1}{n}\sum_{i=1}^{n}|y_i - \hat{y}_i| \tag{21}$$

For each physical target, the better-performing probe was retained for comparative analysis.

This probing framework creates a bridge between the latent AEF representation and conventional remotely sensed observables. Rather than attempting to assign a direct physical label to each dimension point in the AEF embedding space, it asks whether information related to vegetation condition, moisture status, spectral reflectance, soil/background response, SAR structure, and temporal dynamics can be decoded from the embedding. In this sense, the analysis separates predictive explainability, where individual latent dimensions can be attributed to model outputs, from physical interpretability, where the latent representation is examined in relation to measurable EO quantities. The resulting evidence therefore characterizes AEF not as a set of directly interpretable physical variables, but as a latent representation whose physical information content can be quantitatively probed and compared across sensor and biophysical domains.

## 4.7 Methodology

The proposed framework is organized into six sequential steps that transform raw ground truth field observations and multi-source satellite data into validated crop type and irrigation system classifications :

1. Data and Study Area: Field-based reference observations were combined with AlphaEarth embeddings and optical and radar satellite data to establish a common dataset for agricultural land-use/land-cover (LULC) and irrigation system classification.
2. Representation Spaces: Three complementary predictor representations were considered: AlphaEarth embeddings, optical and SAR features, and derived spectral indices. These representations enabled the assessment of different forms of satellite-derived information using consistent reference samples.
3. Preprocessing: The datasets were harmonized and organized into analysis-ready feature matrices, ensuring consistency across the different predictor representations and classification tasks
4. AI Benchmark: XGBoost, CatBoost, and TabPFN were evaluated for LULC and irrigation system classification using independent holdout validation and five-fold cross-validation. Their predictive performance and computational efficiency were compared.
5. Explainability and Driving Factors (XAI): SHAP and native XGBoost feature importance were used to identify influential predictors and examine the relative contributions of latent embedding dimensions and physically interpretable EO variables.
6. Physical Probing of AlphaEarth Embeddings: Linear and nonlinear probing approaches were employed to investigate the extent to which AlphaEarth embeddings encode information associated with observable spectral, vegetation, moisture, soil, SAR, and temporal characteristics.

The benchmarking framework was applied independently to two classification tasks: agricultural LULC and irrigation-type classification, allowing the performance of each feature configuration and classifier to be compared consistently across both agricultural mapping applications. In addition to predictive accuracy, two complementary explainability analyses were conducted: (i) identification of the main predictive drivers by comparing feature-attribution patterns for AlphaEarth embeddings and conventional Earth-observation variables, and (ii) assessment of the physical information encoded in the embeddings through the reconstruction of biophysical descriptors using Spearman correlation and probing models based on Ridge regression and Random Forest. These analyses address three overarching questions: which representation provides the most effective basis for agricultural mapping, which features contribute most strongly to model predictions, and whether AEF latent space can be used to recover Earth-observation properties.

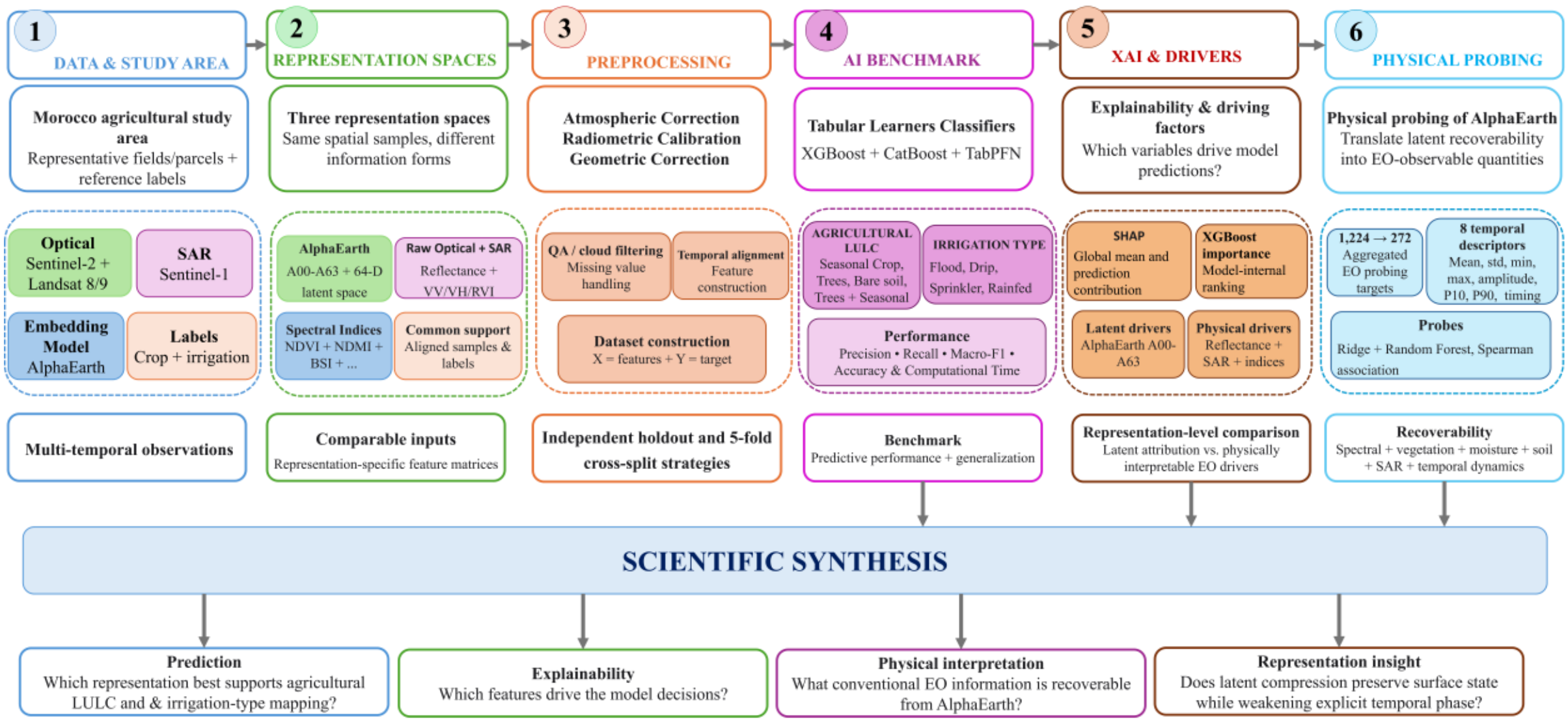

*Figure 2: Integrated methodological framework for benchmarking AlphaEarth embeddings against multi-sensor Earth-observation representations, explaining model drivers, and probing their physical information content for agricultural land-use and irrigation-type mapping in Morocco.*

## 4.8 Computational setup and software environment

All experiments were carried out on a high-performance Linux server, HPC, running Ubuntu 22.04.5 LTS (Jammy) on x86_64 architecture, that is maintained and hosted by the Research Centre for Artificial Intelligence in Geomatics at the Hong Kong Polytechnic University. The system was configured with 24 CPU workers (as indicated by optimal thread allocation for XGBoost) and GPU, NVIDIA A100 80GB PCIe GPUs available for CatBoost and TabPFN via CUDA. All experiments were implemented in Python using libraries including: Data Science Toolkit (El Hachimi et al., 2022), scikit-learn, XGBoost, catboost, and TabPFN libraries.

# 5 Results and Discussion

## 5.1 Agricultural LULC classification

The evaluation of agricultural LULC classification presents distinct challenges relative to irrigation mapping, primarily due to the higher dimensionality of land cover categories (bare soil, trees, trees + seasonal, seasonal crop) and the increased spectral heterogeneity within mixed agricultural systems. Figure 3 presents the comprehensive performance landscape for crop classification.

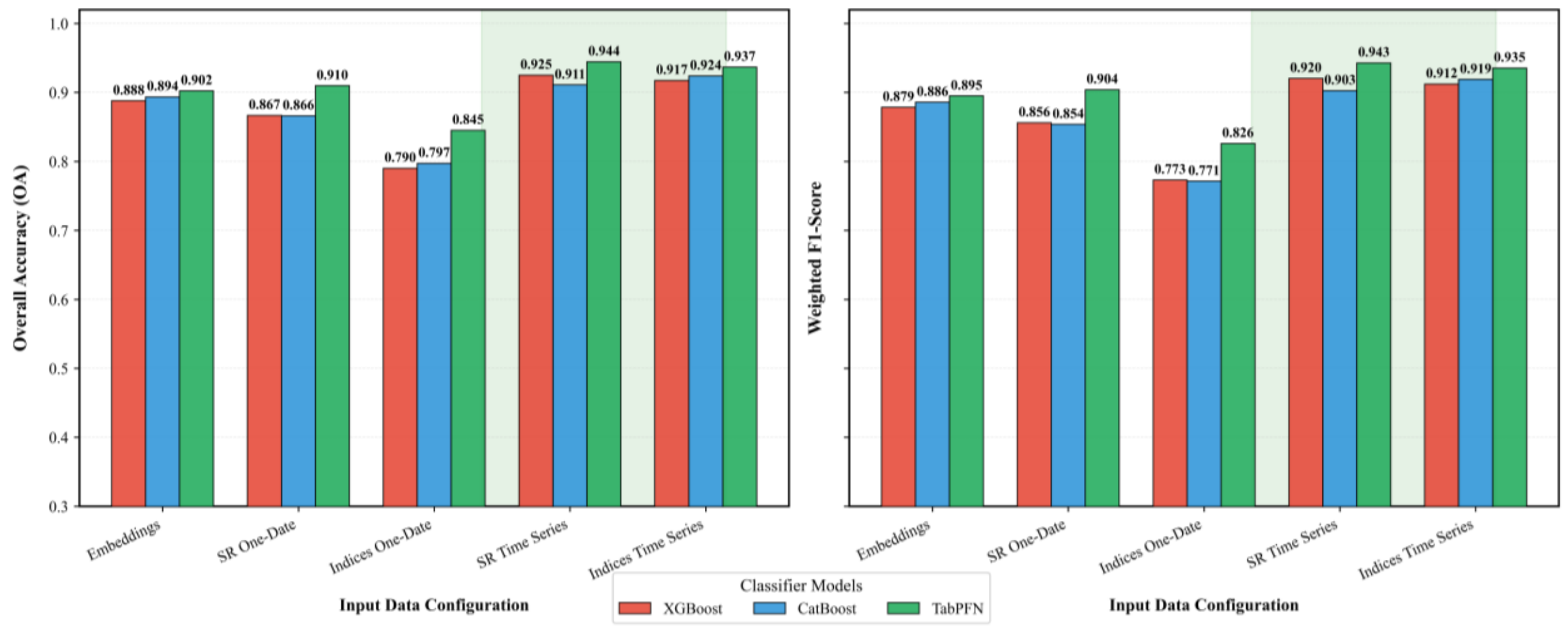

*Figure 3: Crop classification performance comparison across dataset configurations using XGBoost, CatBoost, and TabPFN.*

Across all configurations, temporal data consistently outperformed single-date inputs, with the time series configurations yielding the highest classification accuracies. The Indices Time Series configuration achieved the best overall performance, with TabPFN attaining the highest OA of 0.94 and a Weighted F1-Score of 0.94, followed closely by CatBoost (OA = 0.92; F1 = 0.92) and XGBoost (OA = 0.92; F1 = 0.91). The SR Time Series configuration produced comparable results, with TabPFN again leading (OA = 0.94; F1 = 0.94), suggesting that multi-temporal spectral information provides a strong discriminative signal regardless of whether raw spectral reflectance or derived indices are used.

In contrast, single-date configurations showed markedly reduced performance. The Indices One-Date configuration yielded the lowest accuracies across all models, with OA values of 0.79, 0.80, and 0.85 for XGBoost, CatBoost, and TabPFN, respectively, and corresponding F1-Scores of 0.77, and 0.83. The SR One-Date and Embeddings configurations exhibited intermediate performance, with OA values ranging from 0.87 to 0.91, indicating that pre-computed embeddings capture meaningful spectral-spatial features even without explicit temporal information.

Regarding classifier performance, TabPFN consistently outperformed XGBoost and CatBoost across nearly all input configurations, a finding that aligns with recent studies demonstrating the effectiveness of transformer-based tabular learners in remote sensing classification tasks (Tabasum et al., 2025).

The consistent superiority of time-series configurations over single-date inputs (Figure 3) reinforces the fundamental importance of temporal dynamics in LULC classification. The performance gap between time-series and single-date configurations is substantial, with TabPFN on Indices time series achieving an OA of 0.94 compared to 0.85 on Indices one-date (a relative improvement of 10.9%). This finding aligns with established principles in agricultural remote sensing: crop types and land cover classes exhibit distinct phenological patterns that are best captured through multi-temporal observations (Elwan et al., 2022).

To assess the sensitivity of classifier performance and data configuration to land cover class, we evaluated the F1-score of each model for each crop class individually (Figure 4). This class-specific analysis is essential for identifying which land cover types are reliably classified, which remain challenging, and how different data configurations influence classification accuracy across heterogeneous landscapes.

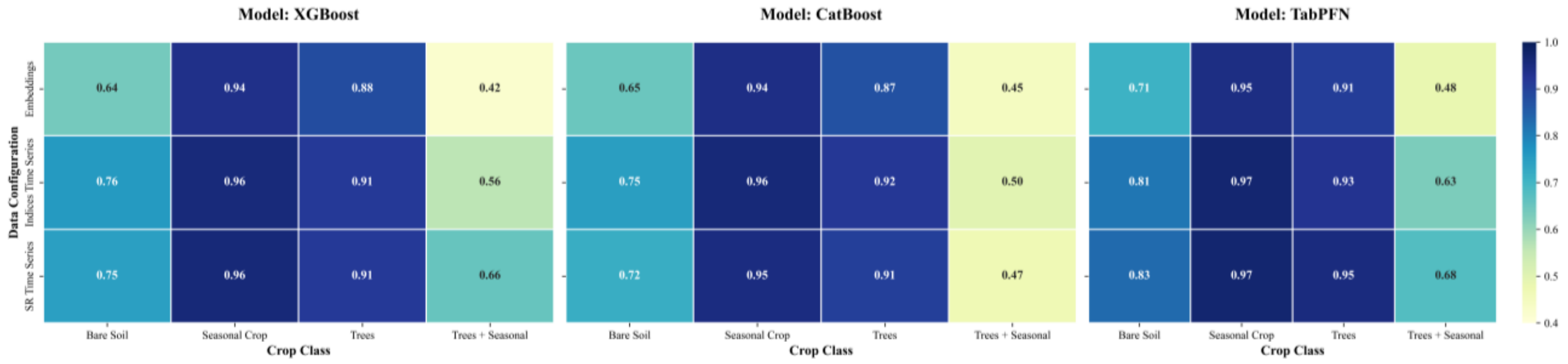

*Figure 4: Classification performance of XGBoost, CatBoost, and TabPFN across different data configurations and crop classes*

Seasonal Crop consistently achieves the highest F1-scores across all modalities and models, ranging from 0.94 (XGBoost, Embeddings) to 0.97 (TabPFN, SR time series). This suggests that seasonal crop types exhibit distinct spectral and phenological signatures that are readily captured by both static embeddings and time-series data. The high performance for seasonal crops is consistent with their well-defined growing cycles and associated agricultural practices, which create strong discriminative signals in remote sensing data.

Trees show strong performance on time-series configurations, with F1-scores reaching 0.95 (TabPFN, SR and Indices time series). However, performance drops significantly under the Embedding configuration, with F1-scores ranging from 0.87 (CatBoost) to 0.91 (TabPFN). The high performance on time-series data reflects the distinct phenological patterns of tree cover, which exhibit clear seasonal cycles that are captured by multi-temporal observations (Modanesi et al., 2025; Zheng et al., 2024). The lower performance under the Embedding configuration suggests that static embeddings may not fully capture the spectral-spatial characteristics of tree classes, particularly in agricultural landscapes where trees may be intermixed with other vegetation types. This underscores the importance of temporal information for accurately mapping woody vegetation in heterogeneous environments.

Trees+Seasonal represents the most challenging class across all configurations, with F1-Scores ranging from 0.42 (XGBoost, Embeddings) to 0.68 (TabPFN, SR time series). The consistently lower performance for this mixed class can be attributed to spectral confusion between tree cover and seasonal agricultural vegetation, particularly during periods of active crop growth. This finding aligns with previous studies highlighting the difficulty of discriminating between woody vegetation and seasonal crops in heterogeneous agricultural landscapes (Zvonkov et al., 2025).

Bare Soil exhibits moderate to good performance, with F1-Scores ranging from 0.64 (XGBoost, Embeddings) to 0.83 (TabPFN, SR time series). The substantial improvement on time-series configurations, particularly for TabPFN, suggests that multi-temporal observations better capture the seasonal dynamics of bare soil areas, which may be affected by agricultural cycles, fallow periods, and climate conditions (Shamal and Weatherhead, 2014). The comparatively lower classification performance observed for Bare Soil and Trees + Seasonal may also be partly associated with the unequal distribution of reference samples among LULC classes. Specifically, Bare Soil accounted for only 942 samples (9.07% of the dataset), while Trees + Seasonal was the least represented class, with 385 samples (3.71%). This imbalance may have limited the model’s ability to adequately learn the spectral and temporal variability of these classes, particularly compared with more frequently represented categories such as Seasonal Crop and Trees. Furthermore, despite the potential of multi-temporal spectral indices to distinguish bare soil from vegetated surfaces, spectral similarities with sparsely vegetated or degraded shrubland and agricultural fields during periods of low vegetation cover may contribute to classification uncertainty.

This result is consistent with recent studies demonstrating the effectiveness of transformer-based architectures for tabular data classification (Hollmann et al., 2023a). TabPFN's self-attention mechanism enables the model to capture complex feature interactions that are not easily modeled by traditional tree-based ensemble methods (Schmidinger et al., 2026). The model's in-context learning capability allows it to perform robust inference from limited labeled samples (Halder et al., 2026), which is particularly valuable for remote sensing applications where ground-truth data are often scarce. The choice of data configuration had a pronounced impact on classification accuracy. Embeddings consistently yielded the lowest performance across all models and classes. This suggests that static, non-temporal representations derived from embeddings are insufficient to capture the phenological dynamics that distinguish different crop types, especially when classes share similar spectral signatures. In contrast, SR and Indices time series significantly improved classification performance, with F1-scores exceeding 0.90 for most classes when using TabPFN. The temporal information encoded in these configurations allows the models to leverage seasonal variations in vegetation indices (e.g., NDVI, NDMI) and surface reflectance, which are critical for distinguishing between crop types that may appear spectrally similar at a single point in time (Morell-Monzó et al., 2023). The slight advantage of SR time series over Indices time series suggests that raw surface reflectance provides complementary information to derived indices, particularly for capturing subtle variations in canopy structure and soil background (Yousaf et al., 2025).
The model's robustness was evaluated using 5-fold cross-validation across three data configurations (LULC embeddings, surface reflectance time series, and spectral indices). The F1-score was used as the primary performance metric to account for inherent class imbalances in the LULC dataset (Figure 5).

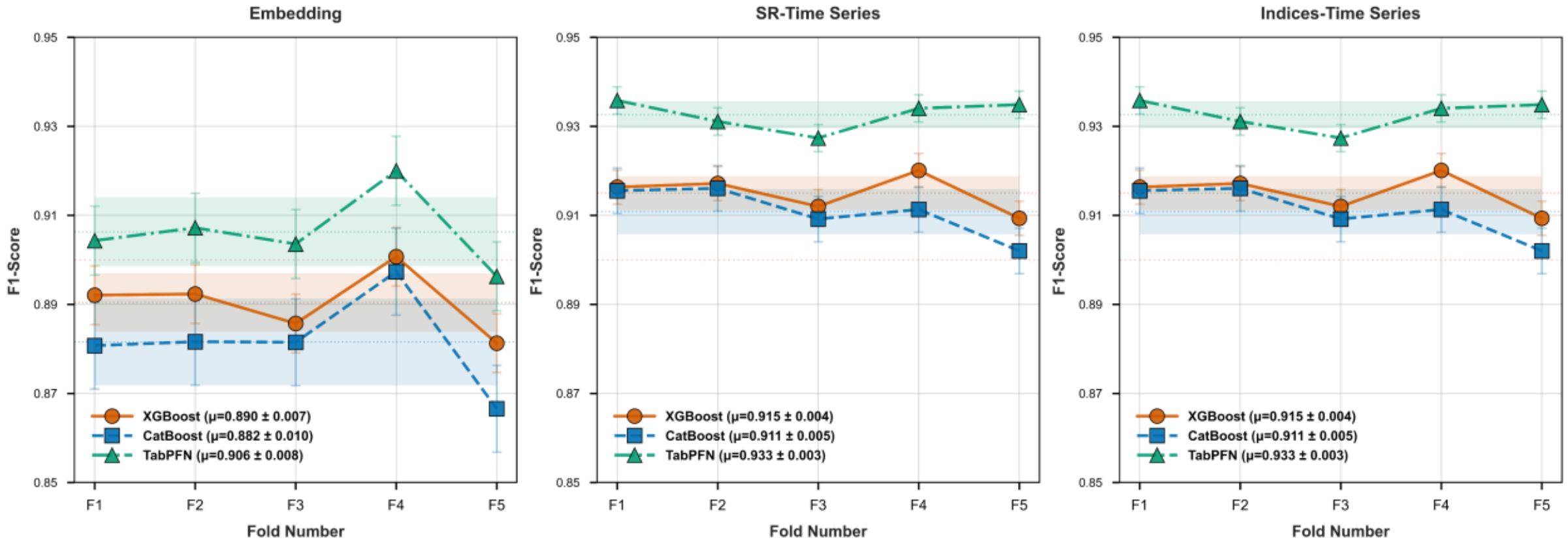

*Figure 5: F1-score performance across 5-fold cross-validation for three machine learning models (XGBoost, CatBoost, TabPFN) under three feature configurations: (a) LULC Embedding, (b) SR-time series, and (c) Indices-time series.*

Figure 5 presents the stability of machine learning models across different training-validation splits for all model-configuration combinations. TabPFN exhibited superior robustness compared to the tree-based ensembles, maintaining the lowest inter-fold variability across all configurations. For the multi-temporal configurations (Surface Reflectance and spectral indices), TabPFN achieved a remarkably low standard deviation (SD) of 0.003 around a mean F1-score of 0.93. While variability increased slightly for the Embedding configuration (SD = 0.009), TabPFN remained the most consistent model. This combination of high predictive accuracy and low variance suggests that TabPFN's transformer-based architecture, which leverages self-attention to model feature interactions, is inherently less sensitive to minor fluctuations in the training data distribution than traditional methods.
In contrast, XGBoost and CatBoost demonstrated higher sensitivity to the composition of the CV folds. XGBoost showed moderate variability, with SDs ranging from 0.004 to 0.007, while CatBoost exhibited the highest fluctuations (SD: 0.005-0.010). These tree-based models showed pronounced performance dips in specific iterations, most notably in Fold 5, where CatBoost's F1-score dropped to 0.87 in the Embedding configuration. The increased variability in these models can be attributed to the nature of recursive partitioning and sequential boosting; these mechanisms can overfit to specific characteristics of a training subset, especially when certain land cover classes are underrepresented or when temporal signals are absent.

The stability advantage of TabPFN has important implications for operational LULC classification. The model's low inter-fold variability ensures consistent performance across different geographic regions, acquisition dates, and training data compositions, reducing the need for frequent model retraining (Halder et al., 2026). This reliability, combined with superior accuracy on time-series data (F1 of 0.93), makes TabPFN the recommended choice for operational deployment where consistency and accuracy are equally critical.

### Spatial extent of LULC mapping

To evaluate the spatial applicability of three configurations for land-use and land-cover (LULC) mapping, a continuous Area of Interest (AOI) covering approximately 1,000 km² was selected (Figure 1).

The selected AOI was subsequently used as a common spatial framework to generate the 2024 land-use/land-cover (LULC) maps using three feature configurations: satellite-derived surface reflectance data, AlphaEarth satellite embeddings, and satellite-derive spectral indices. Although the broader study covered the 2023-2025 period, 2024 was selected for detailed spatial mapping primarily because most of the field reference data were collected during this year, ensuring greater temporal consistency between ground observations and satellite-derived predictors. Furthermore, 2024 represents the central year of the study period, providing a consistent temporal reference for comparing the different feature representations. The AOI selection was guided by the spatial distribution of the available reference polygons, to include all four LULC classes and improve their representation within a single continuous geographical extent. The selected AOI contained 729 reference polygons, comprising 80 bare soil, 416 seasonal crop, 113 tree, and 120 mixed trees and seasonal crop polygons. By maintaining the same geographical extent and reference year across the three configurations, the analysis facilitated a consistent comparison of classification accuracy, spatial distribution patterns, and estimated LULC class areas. Figure 9 presents the LULC classification maps generated from the three feature configurations. The classification outputs illustrate the distribution of bare soil, seasonal crops, trees, and mixed trees + seasonal crop systems across the AOI.

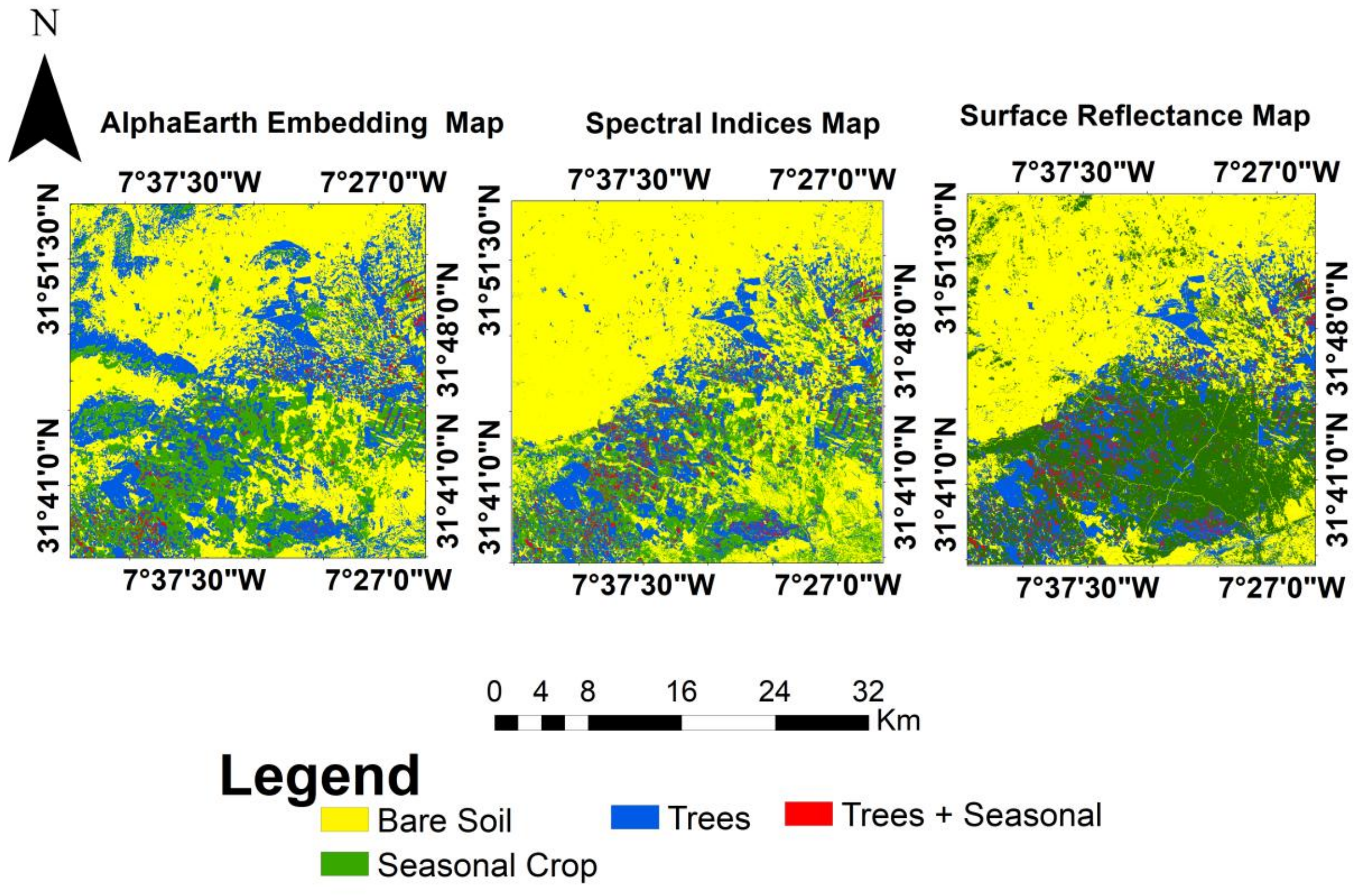

*Figure 6: The LULC maps of three feature configurations: AlphaEarth embedding, surface reflactance, and spectral indices.*

Figure 6 presents the spatial distribution of four land-use/land-cover (LULC) classes (bare soil, seasonal crops, trees, and trees + seasonal crops) generated using the TabPFN classifier with three feature configurations: AlphaEarth satellite embeddings, spectral indices, and surface reflectance. All three maps cover the same geographical extent and use an identical classification scheme, allowing direct comparison of the resulting spatial patterns.

Visual inspection reveals notable differences in the spatial distribution and fragmentation of LULC classes among the three configurations. Bare soil dominates the northern portion of the study area across all three maps, whereas the central and southern sectors exhibit a more heterogeneous agricultural landscape composed of seasonal crops, trees, and mixed vegetation. The AlphaEarth embedding-based classification displays a particularly heterogeneous spatial pattern, with extensive tree-covered areas interspersed with seasonal crops and bare soil. In contrast, the spectral-indices-based classification shows a more continuous distribution of bare soil in the northern sector, while tree-covered and agricultural areas are concentrated mainly in the central and southern regions. The surface-reflectance-based map also identifies bare soil as the dominant class in the northern sector, but exhibits a more extensive concentration of seasonal crops in the central and southern portions.

The quantitative LULC area estimates further highlight these differences (Table 6). Bare soil represents the largest proportion of the classified landscape in all configurations, covering 490.26 km² (49.00%) using AlphaEarth embeddings, 608.60 km² (60.79%) using spectral indices, and 596.22 km² (59.59%) using surface reflectance. The most pronounced difference concerns the Trees class, which occupies 298.06 km² (29.79%) in the AlphaEarth classification, compared with 181.80 km² (18.09%) for spectral indices and 145.82 km² (14.57%) for surface reflectance. Conversely, the Seasonal Crop class is more extensive in the surface-reflectance classification (236.54 km²; 23.64%) than in the AlphaEarth (193.36 km²; 19.33%) and spectral-indices (182.04 km²; 18.18%) classifications. The Trees + Seasonal class occupies a comparatively small proportion of the study area across all configurations, ranging from 1.88% for AlphaEarth to 2.93% for spectral indices.

Despite these substantial differences in mapped class proportions, the three configurations achieve comparable classification performance. Surface reflectance and spectral indices yield an overall accuracy of 0.86, compared with 0.85 for AlphaEarth embeddings. Surface reflectance achieves the highest balanced accuracy (0.82), followed by spectral indices (0.80) and AlphaEarth embeddings (0.79). The Kappa coefficients vary only slightly, from 0.76 for AlphaEarth and surface reflectance to 0.77 for spectral indices, while the reported F1-scores range from 0.811 to 0.82.

The confusion matrices provide further insight into class-specific classification performance across the three feature configurations. AlphaEarth embeddings achieved the highest recall for seasonal crop (95.20%) and trees (88.24%), compared with 94.40% and 85.29% for spectral indices, and 91.20% and 85.29% for surface reflectance, respectively. However, AlphaEarth exhibited lower recall for Bare Soil (75.00%) and particularly trees + seasonal (58.33%). The spectral-indices configuration achieved comparable recall for bare soil (75.00%) and seasonal crop (94.40%), while providing intermediate performance for Trees + Seasonal (66.67%), outperforming AlphaEarth but remaining below surface reflectance. In contrast, surface reflectance achieved the highest recall for bare soil (79.17%) and trees + seasonal (72.22%), resulting in the highest balanced accuracy (0.82), followed by spectral indices (0.80) and AlphaEarth embeddings (0.79). These results suggest that the three feature representations differ in their ability to discriminate between relatively distinct land-cover categories and heterogeneous agricultural classes. AlphaEarth embeddings showed a slight advantage in identifying Seasonal Crop and Trees, whereas spectral indices maintained competitive performance across the vegetation classes, potentially reflecting their sensitivity to vegetation characteristics. Surface reflectance, meanwhile, demonstrated a more balanced classification performance, particularly for the mixed trees + seasonal class, suggesting that retaining individual spectral-band information may help distinguish heterogeneous vegetation conditions. Nevertheless, these interpretations remain hypotheses concerning the underlying feature characteristics, and recall differences alone do not establish which configuration produces the most reliable spatial distribution or class-area estimates.

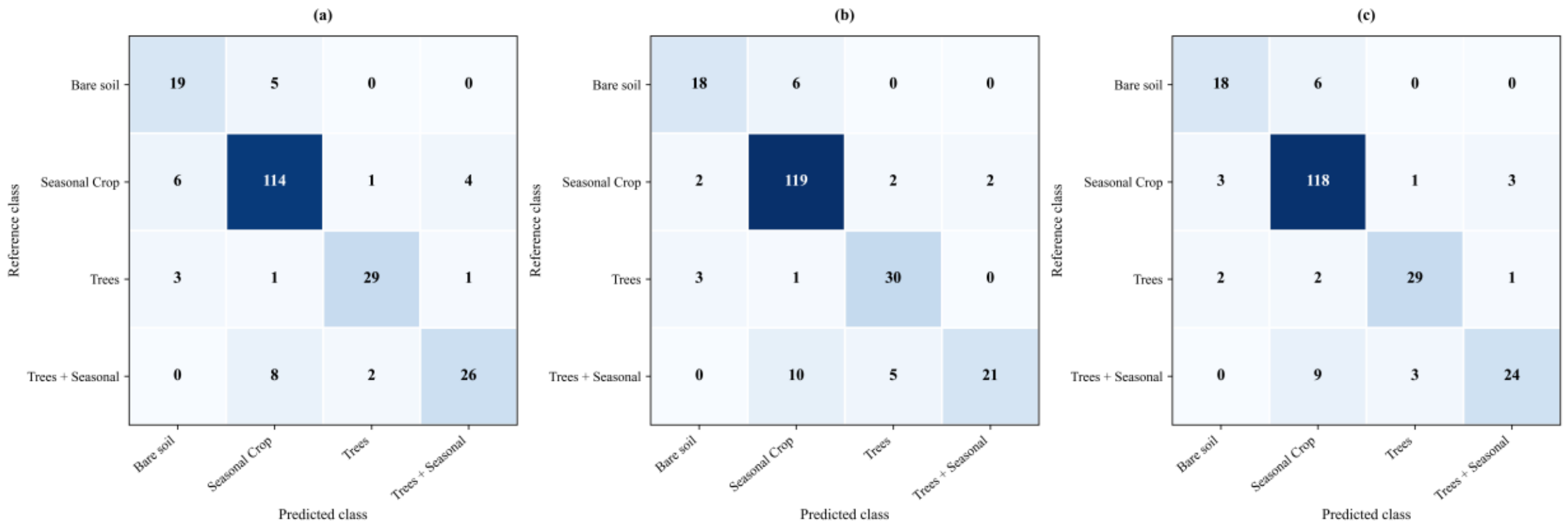

*Figure 7: Confusion matrices of TabPFN based LULC classification across three satellite-derived predictor configurations. (a) Surface reflectance, (b) AlphaEarth embeddings, and (c) Spectral indices.*

*Table 6. Comparison of land-use/land-cover (LULC) area distributions and classification accuracy across three satellite feature configurations in 2024.*

| Configurations | **Surface Reflectances** | | **AlphaEarth Embedding** | | **Spectral Indices** | |
|---|---|---|---|---|---|---|
| Classes | Area ($km^2$) | Percentage | Area ($km^2$) | Percentage | Area ($km^2$) | Percentage |
| Bare soil | 596.22 | 59.59% | 490.26 | 49.00% | 608.60 | 60.79% |
| Seasonal Crop | 236.54 | 23.64% | 193.36 | 19.33% | 182.04 | 18.18% |
| Trees | 145.82 | 14.57% | 298.06 | 29.79% | 181.8 | 18.09% |
| Trees + Seasonal | 21.87 | 2.19% | 18.78 | 1.88% | 29.37 | 2.93% |
| OA | 0.86 | | 0.85 | | 0.86 | |
| Balanced_Accuracy | 0.82 | | 0.79 | | 0.80 | |
| Kappa | 0.76 | | 0.76 | | 0.77 | |
| F1_Score | 0.82 | | 0.81 | | 0.82 | |

Overall, the comparison demonstrates that the three feature configurations achieved comparable classification performance, with overall accuracies ranging from approximately 85% to 86%, despite substantial differences in the estimated spatial distribution of LULC classes. Surface reflectance provided the highest balanced accuracy (0.82), while spectral indices achieved a slightly higher Kappa coefficient (0.77). AlphaEarth satellite embeddings produced a considerably larger mapped area of tree cover (29.79%) than surface reflectance (14.57%) and spectral indices (18.09%), highlighting the influence of feature representation on class-area estimates. These findings indicate that similar global accuracy metrics do not necessarily imply equivalent spatial representations of agricultural landscapes, consistent with the observations of (Stehman and Foody, 2019). Furthermore, as emphasized by (Olofsson et al., 2014), reliable land-cover area estimation requires rigorous accuracy assessment and appropriate reference data. Consequently, the selection of satellite-derived features should consider not only overall classification accuracy but also class-specific performance and the reliability of spatial area estimates.

The agricultural LULC and irrigation practices provides an important link between the two classification tasks. Irrigation systems are often associated with specific crop types and vegetation structures, with drip irrigation commonly used in orchards and perennial crops, while flood irrigation is frequently associated with seasonal crops. These relationships suggest that the spectral, structural, and temporal characteristics used to distinguish agricultural LULC classes may also contribute to irrigation-type discrimination. However, irrigation practices are not exclusively determined by crop type, and similar crops may be managed under different irrigation systems. Therefore, following the evaluation of LULC classification, we further investigated the ability of AlphaEarth embeddings and conventional remote-sensing features to discriminate irrigation types, thereby assessing whether these representations capture information related to agricultural water management beyond land-cover characteristics alone.

## 5.2 Irrigation type discrimination

The performance evaluation of irrigation method classification reveals substantial variation across models and feature representations (Figure 8).

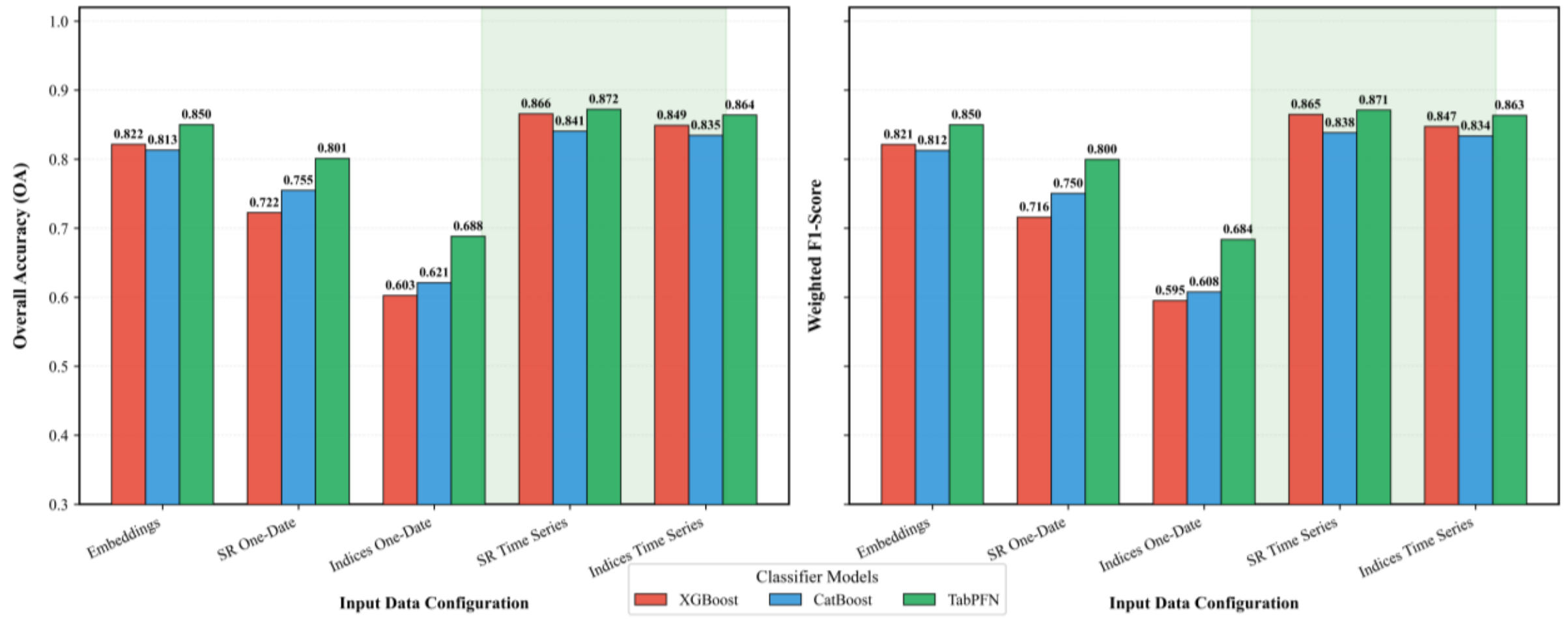

*Figure 8: Leave one out validation results for irrigation classification performance comparison across dataset configurations.*

Figure 8 presents a comprehensive comparison of overall accuracy (OA) and F1-score for three classifiers (XGBoost, CatBoost, and TabPFN) evaluated across five distinct dataset configurations. The figure reveals several critical patterns. First, TabPFN achieves the highest LOOV accuracy and F1-Score consistently outperforms both gradient boosting frameworks (XGBoost and CatBoost) through all dataset configurations, with performance advantages most pronounced in time series configurations (SR time series and indices time series), as highlighted by shaded region. Second, the minimal discrepancy between accuracy and weighted F1-Score ($\Delta < 0.005$ for most configurations) indicates relatively balanced class distributions in the irrigation dataset, with no severe imbalance penalizing minority classes (Hoque et al., 2025). Third, the dramatic performance degradation observed in single-date configurations (F1 < 0.70 for all models) underscores the critical importance of raw spectral information over derived indices for irrigation discrimination, likely because spectral indices compress subtle reflectance variations associated with different water application methods (Chen et al., 2025).

A critical outcome of this benchmark is the near-parity observed between the AlphaEarth neural embeddings and the explicit multi-temporal time-series datasets. While the SR time series configuration achieved the highest accuracy (OA = 0.87), it outperformed the AlphaEarth embeddings, which reached a remarkable performance with an OA of 0.85 and an F1-score of 0.85. While (Yang et al., 2026) demonstrated the potential of AEF embeddings for irrigation mapping, they also acknowledged that the model's performance is context-dependent and that substantial degradation was observed under cross-region transfer. The approximately 2.2% accuracy difference indicates that the self-supervised foundational model has learned much of the essential spatiotemporal structure relevant to irrigation type discrimination.

While overall accuracy metrics offer a consolidated measure of model performance, they frequently mask critical disparities in class-specific discriminative capacity. To elucidate these underlying patterns and assess the thematic fidelity of each type irrigations, we decomposed the classification outcomes into per-class precision, recall, and F1-score across the four principal irrigation modalities: drip irrigation, rainfed, sprinkler irrigation, and flood irrigation (Figure 9).

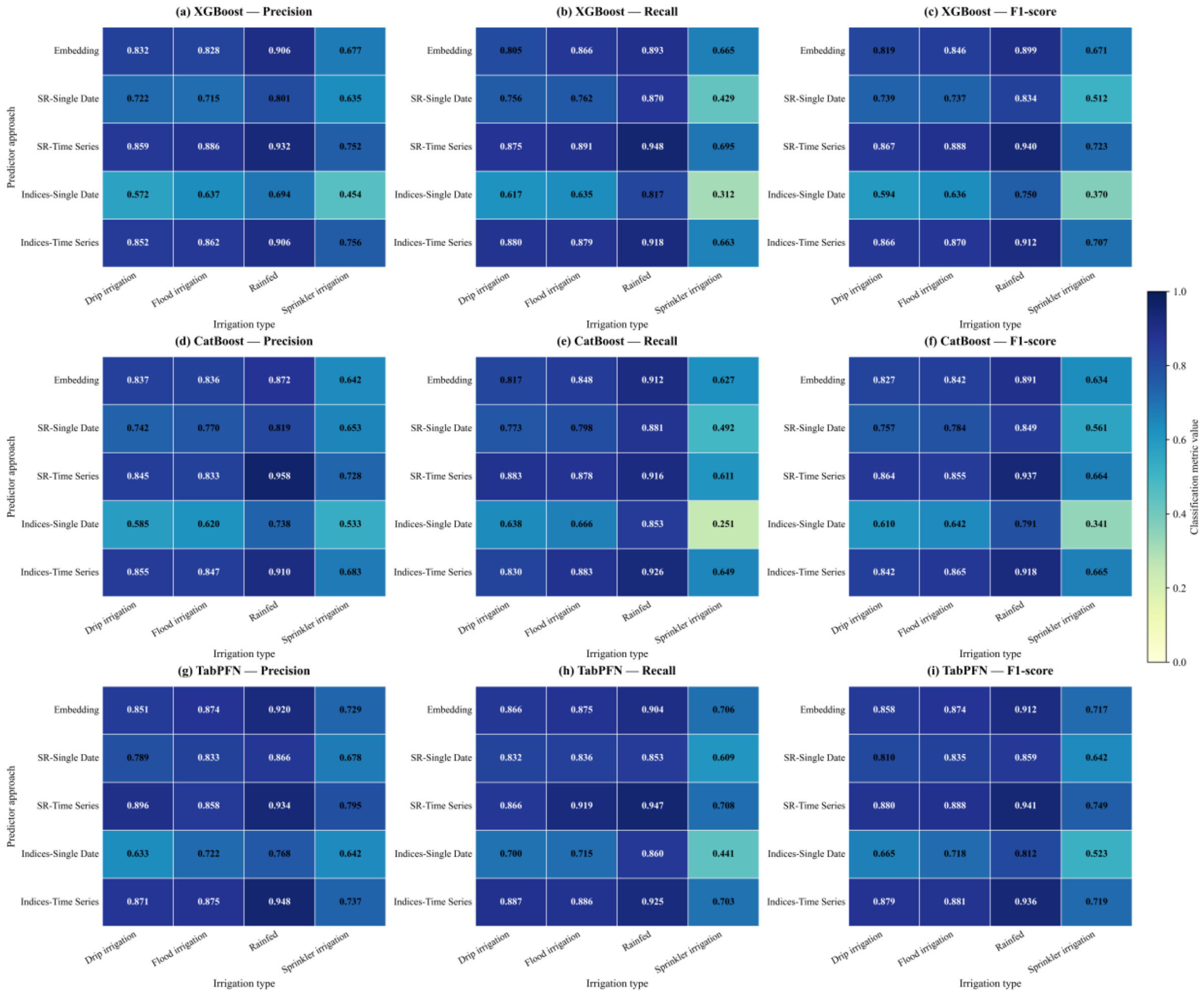

*Figure 9: Per-class performance matrix (Precision, Recall, and F1-Score) for irrigation classification.*

The classification results revealed clear differences in the ability of the investigated satellite representations to identify rainfed agriculture and distinguish among irrigation systems. Rainfed agriculture was generally well classified using the multi-temporal configurations, with F1-scores ranging from 0.89 to 0.94 across the three machine learning models. However, classification performance declined when single-date observations were used. For single-date surface reflectance (SR), rainfed F1-scores ranged from 0.83 to 0.86, whereas single-date spectral indices produced lower values ranging from 0.75 to 0.81. In comparison, SR time series achieved F1-scores close to 0.94, while spectral-index time series yielded values between 0.91 and 0.93. (Shamal and Weatherhead, 2014) demonstrated that rainfed exhibits distinct spectral signatures because these systems are primarily driven by natural precipitation patterns, which create consistent and predictable vegetation responses that are more readily discriminated from other land cover types.

In contrast, the lower performance for sprinkler irrigation, ranging from 0.63 to 0.75 can be attributed to several factors. First, sprinkler-irrigated and rainfed fields may show similar spectral characteristics when natural precipitation is sufficient to satisfy crop water requirements. Under these conditions, differences in vegetation water status and growth between irrigated and rainfed crops may become less pronounced, reducing their spectral separability and potentially increasing classification uncertainty. Particularly when rainfall events reduce differences in water stress and chlorophyll content. This spectral similarity likely accounts for the confusion between sprinkler-irrigated and rainfed classes in our classification results. Second, the operational variability of sprinkler systems, including differences in sprinkler type (impact, gear, reaction), application rates, and irrigation scheduling, contributes to within-class spectral heterogeneity (Nouwakpo et al., 2024).

Among the three input configurations, the surface reflectance and indices time series consistently outperformed the Embedding configuration, particularly for the rainfed class, where TabPFN achieved the highest F1-score (0.93). TabPFN also demonstrated the strongest overall performance, attaining the highest F1-score (0.75) for the most challenging class (sprinkler irrigation) using the surface reflectance time series. The superior performance of time-series-based configurations highlights the importance of capturing irrigation-related phenological dynamics throughout the growing season. Studies demonstrated that multi-temporal feature fusion frameworks achieve overall accuracies exceeding 90% for irrigation identification (Li et al., 2025). The study conducted by (Gao et al., 2018), demonstrated that SAR time-series metrics, including mean backscatter, variance, correlation length, and fractal dimension, achieved good overall accuracy (81.08%) for irrigation mapping.

To evaluate the generalization capacity of each model configuration combination and ensure that the reported performance metrics were not artifacts of a particular train-test partition, we conducted a stratified 5-fold cross-validation (CV5) protocol. Figure 5 presents the per-fold trajectories of accuracy, precision, recall, and F1-score across the three input configurations (embedding, SR-time series, and Indices time series, with model-specific mean values (μ) annotated for direct comparative reference. This disaggregated visualization serves a dual purpose: it validates the robustness of the optimal configurations identified in the preceding analyses and quantifies the sensitivity of each classifier to sample variability inherent in the dataset.

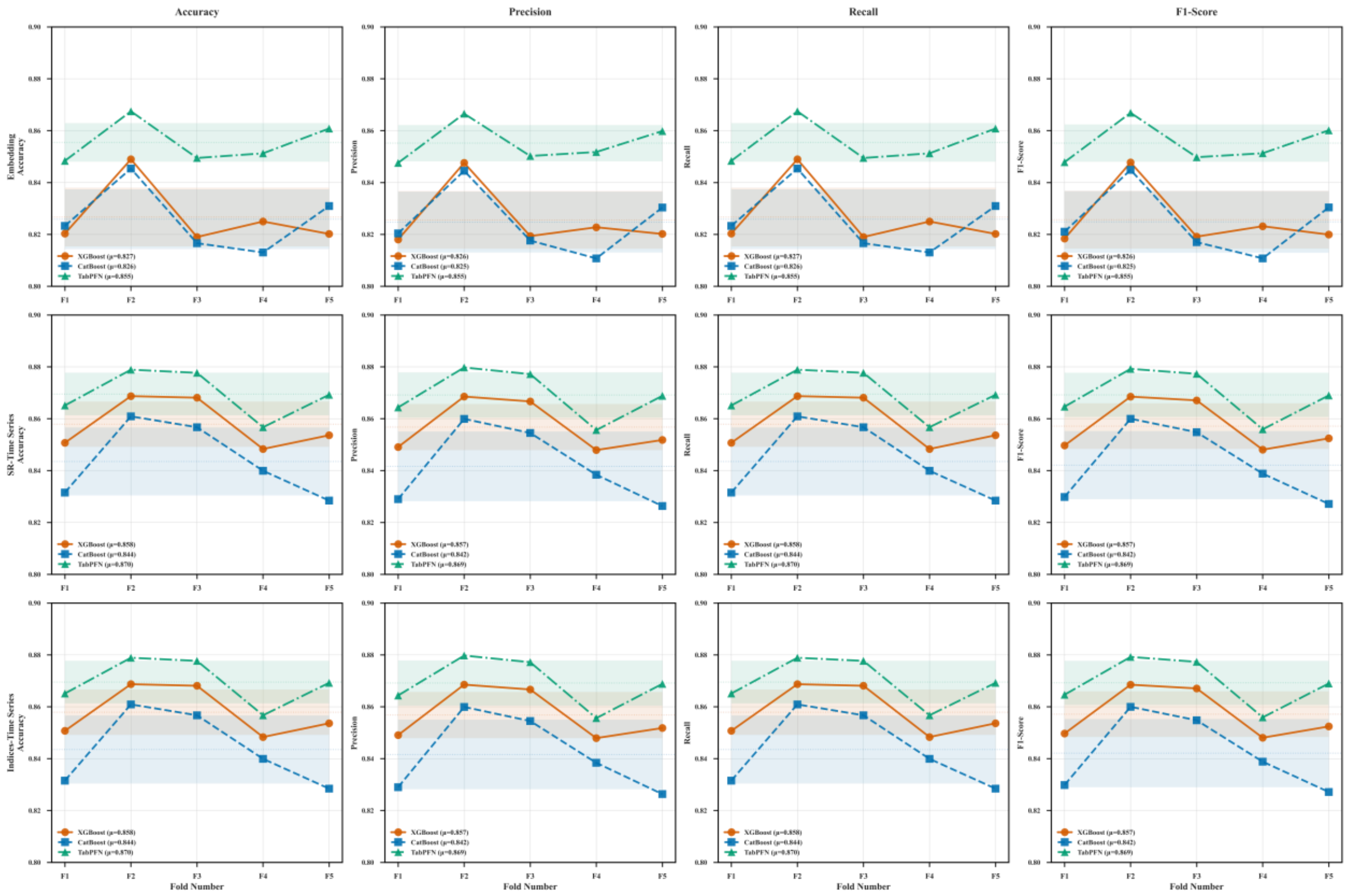

*Figure 10: 5-fold cross-validation performance trajectories stratified by input configuration and evaluation metric.*

The 5-fold cross-validation results in Figure 10 reveal consistent performance differences across the three data configurations and machine learning models. Across all models and metrics, the time-series configurations (SR and Indices) substantially outperform the Embedding configuration.

Under the Embedding configuration, all three models show noticeable fold-to-fold variation, with accuracy swinging by up to 0.03 between consecutive folds. Within this volatile landscape, TabPFN stands out as the clear exception. Its performance remains remarkably steady, hovering within ±0.012 of a mean of 0.86 across all four evaluation metrics. The last three folds (F3-F5) are especially telling: accuracy varies by only 0.01 (ranging from 0.85 to 0.86), and the gap between precision and recall stays well below 0.008. In contrast, XGBoost and CatBoost cluster around a lower mean of 0.83 with considerably more fluctuation. Meaning TabPFN is roughly three times more stable despite working from identical inputs. The consistent performance of TabPFN across all configurations suggests that transformer-based architectures are well-suited for irrigation classification tasks. The model's ability to maintain high accuracy even with limited training samples makes it particularly attractive for regions where ground-truth data are scarce.

The precision and recall observed across models reveal important class-specific behaviors. For the Embedding configuration, XGBoost achieves higher precision (0.83 ± 0.01) than recall (0.83 ± 0.01), suggesting a tendency toward conservative predictions that minimize false positives. In contrast, CatBoost and TabPFN show more balanced precision-recall profiles, indicating better overall class discrimination.

TabPFN under the Embedding configuration reflects a meaningful synergy between the globally consistent representations produced by AlphaEarth's self-supervised pretraining and TabPFN's in-context learning mechanism, which together appear to absorb the distributional shifts between folds that tend to destabilize gradient-boosted approaches (Bihui Lu et al., 2026).

## 5.3 Training time analysis

The relationship between feature configuration, computational demand, and classification accuracy for TabPFN is summarized in Table 7. Our results reveal a significant trade-off between processing time and predictive performance across both the irrigation and LULC classification targets.

*Table 7. LOOV training time (seconds) and overall accuracy for TabPFN, classification target, and feature configuration.*

| Target | Feature configuration | Time (s) | Accuracy. |
|---|---|---|---|
| Irrigation | Embedding | 14.26 | 85.0% |
| | Surface reflectance (time series) | 2640.94 | 87.2% |
| | Spectral indices (time series) | 2631.18 | 86.4% |
| LULC | Embedding | 277.27 | 90.2% |
| | Surface reflectance (time series) | 4845.87 | 94.4% |
| | Spectral indices (time series) | 2474.91 | 93.7% |

Table 7 indicates a clear trade-off between predictive performance and computational cost across the tested feature configurations. The embedding representation was by far the most computationally efficient, requiring only 14.26 s for irrigation and 277.27 s for LULC classification, but it also produced slightly lower accuracies, with 85.0% for irrigation and 90.2% for LULC. This suggests that although embeddings provide a compact and fast feature representation, they may not fully capture the temporal detail required for optimal classification performance.

In contrast, the surface reflectance time-series configuration achieved the highest accuracies for both tasks, reaching 87.2% for irrigation and 94.4% for LULC. However, this gain in accuracy came at a substantial computational cost, with processing times increasing to 2640.94 s for Irrigation and 4845.87 s for LULC. This represents a nearly 20-fold increase in processing time compared to the Embedding approach for the LULC task, highlighting the intensive nature of transformer-based attention when processing high-dimensional multi-temporal raw bands.

The spectral indices time-series configuration offered an intermediate solution. For LULC, it achieved an accuracy of 93.7%, only 0.7 percentage points lower than surface reflectance, while reducing processing time by nearly 50% (2474.91 seconds). This indicates that spectral indices preserve much of the relevant phenological information while lowering dimensionality and improving computational efficiency. Overall, The results indicate that the embedding approach achieves acceptable accuracies in both tasks, with only limited differences compared with the time-series configurations. Although surface reflectance and spectral indices slightly improve classification performance about 2% in irrigation and 4% for LULC tasks, the gains are modest relative to the substantial increase in processing time. This suggests that embedding offers a good balance between accuracy and computational efficiency, making it a practical and scalable alternative for both irrigation and LULC mapping (Zheng et al., 2025).

The preceding sections have reported detailed results for each data configuration and target variable, showing that time-series representations consistently achieve higher classification accuracies than embedding-based features. Nevertheless, operational deployment must consider computational efficiency alongside accuracy, and these two criteria are often in tension. Figure 11 summarizes the main performance indicators for LULC and irrigation type classification, highlighting the accuracy–efficiency balance across the three feature configurations (Embedding, SR time series, and Indices time series) obtained with TabPFN.

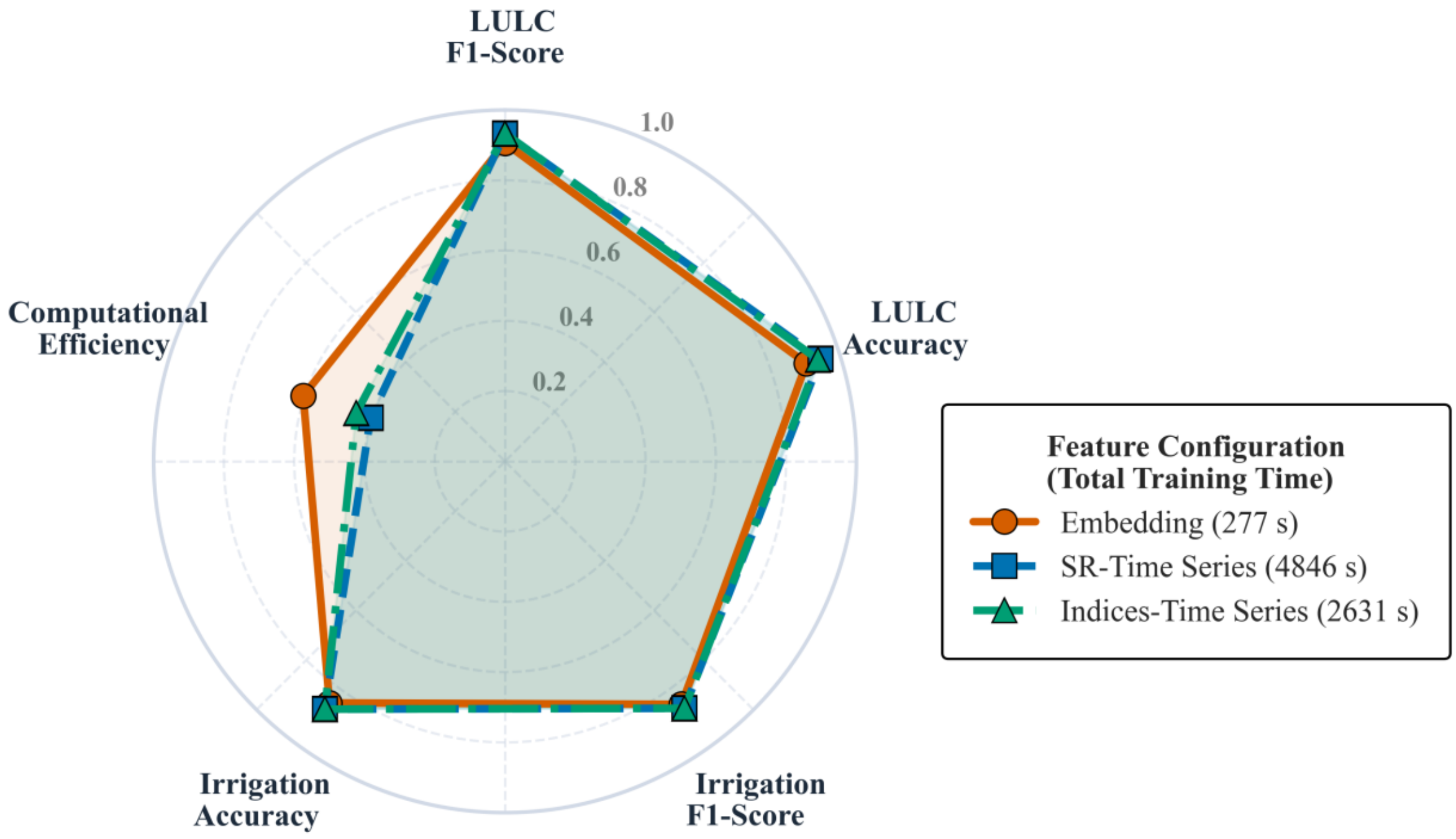

*Figure 11: Performance profile of TabPFN across three feature configurations (Embedding, SR-Time Series, Indices-Time Series) for LULC and Irrigation classification under LOOV.*

The radar chart presented in Figure 11 provides a comprehensive visualization of the performance and computational efficiency for the three feature configurations (Embedding, SR-Time Series, and Indices-Time Series) evaluated with the TabPFN model. The figure integrates five dimensions: LULC F1-Score, LULC Accuracy, Irrigation F1-Score, Irrigation Accuracy, and Computational Efficiency, allowing for a multi-criteria comparison of the different input representations.

The accuracy-efficiency relationship is clearly depicted in the radar chart, which shows that time-series configurations achieve higher classification performance, with improvements of approximately 2.2-3.0%, but require substantially longer training times, by a factor of about 9.5-17.5. This pattern has important practical implications for operational deployment, particularly in settings where computational resources and update frequency must be considered jointly (Xiao et al., 2025).

The practical value of embeddings becomes evident in this context. Although the embedding configuration attains slightly lower accuracy than the time-series alternatives, it offers several operational advantages. First, the markedly reduced training time (277 seconds compared to 2,631-4,846 seconds) makes the embedding-based approach well suited to resource-constrained environments and large-scale applications where computational capacity is limited (Zvonkov et al., 2025). Second, the embedding representation removes the need for extensive time-series preprocessing, thereby simplifying the workflow and facilitating more straightforward integration into existing monitoring pipelines (Brown et al., 2025; Ma et al., 2026a). Third, the compact nature of the embeddings reduces data storage requirements, which is particularly beneficial for long-term, large-area monitoring systems that must handle substantial volumes of input data (Ball et al., 2026; Lu et al., 2025). Together, these characteristics indicate that the embedding configuration provides a pragmatic balance between efficiency and accuracy for operational LULC and irrigation type discrimination.

## 5.4 Feature attribution reveals different predictive drivers across EO representations

The explainability analysis demonstrated that the three feature representations relied on distinct information sources for crop and irrigation classification. The attribution patterns showed a clear contrast between the latent AlphaEarth representation and conventional EO variables. While AlphaEarth embeddings achieved competitive predictive performance using abstract latent dimensions, raw optical, SAR, and spectral-index representations provided direct links to interpretable surface properties such as vegetation condition, moisture availability, and radar-observed canopy structure.

For crop mapping, the AlphaEarth model attribution was dominated by a limited number of latent dimensions, indicating that the foundation-model representation concentrated relevant agricultural information into a subset of the embedding space. According to both SHAP and XGBoost rankings, the most influential variables were latent dimensions such as A56, A54, and A52. However, these dimensions lack predefined physical meanings, which complicates direct interpretation of their environmental significance. Nevertheless, the strong reliance on these latent features suggests that AlphaEarth embeddings effectively encode complex biophysical and phenological patterns that are not fully captured by handcrafted spectral indices or individual optical/SAR bands, consistent with findings from recent studies on geospatial foundation models (Rahman, 2026).

In contrast, the Raw Optical + SAR representation identified physically meaningful drivers. The dominant variables corresponded primarily to spectral bands and radar measurements associated with crop canopy development, biomass structure, and moisture conditions. Variables from the near-infrared and shortwave infrared regions showed strong contributions, consistent with their established sensitivity to vegetation vigor and water content. Sentinel-1 variables, particularly VH-related features, also contributed substantially, suggesting that radar information provided complementary structural information beyond optical observations.

The spectral-index representation highlighted vegetation and moisture-sensitive indices, where NDVI-, NDMI-, and soil-adjusted vegetation metrics contributed strongly to crop discrimination. These results confirm that crop classification benefits from both vegetation phenology and moisture-related information.

*Table 8: XGBoost-based feature importance ranking for crop classification across different Earth-observation representations.*

| | AlphaEarth | Raw Optical + SAR | Spectral Indices |
|---|---|---|---|
| Feature Importance | 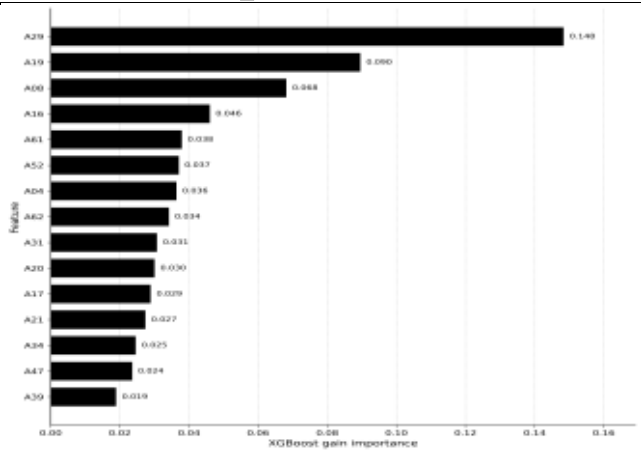 | 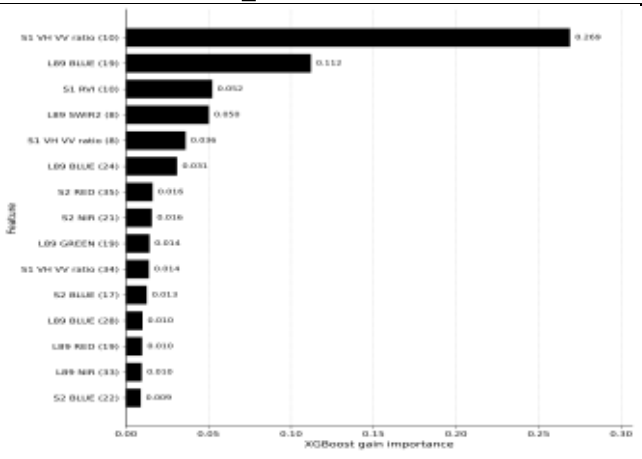 | 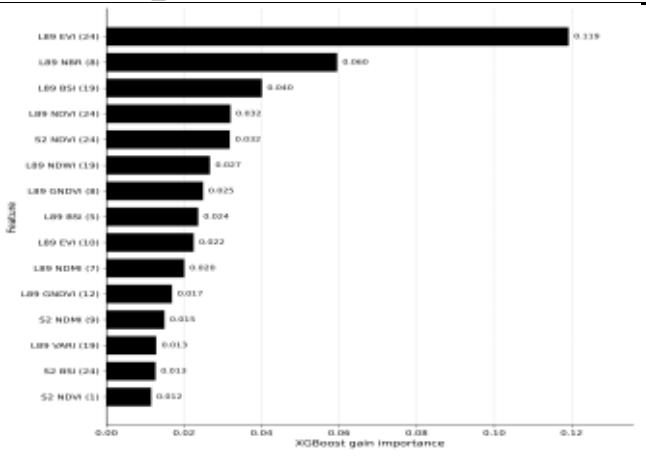 |
| SHAP | 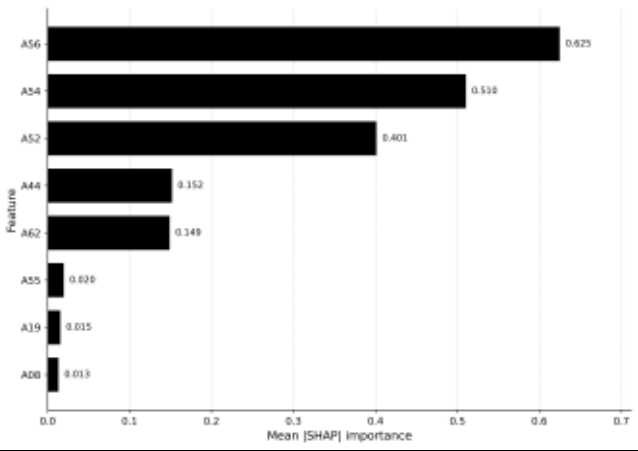 | 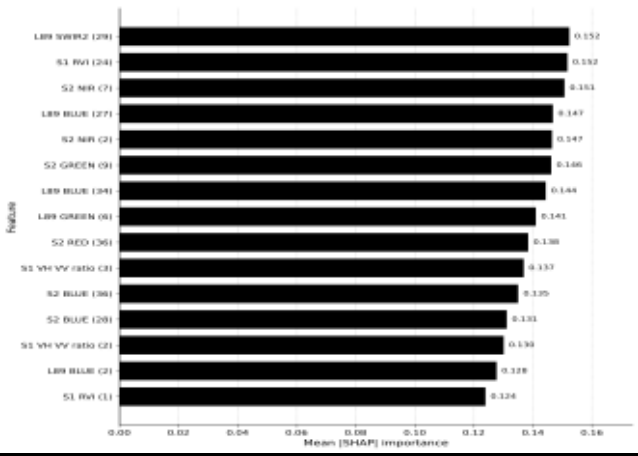 | 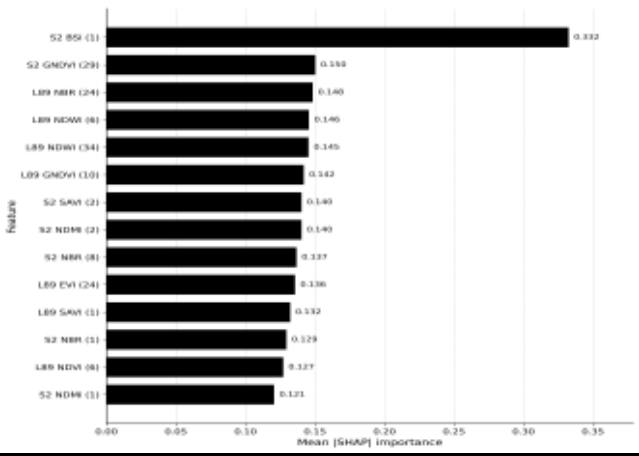 |

For irrigation mapping, the explainability patterns showed stronger contributions from variables associated with vegetation water status and radar response. The AlphaEarth representation again relied on a subset of latent dimensions, with A13 (gain: 0.12, SHAP: 0.10), A47 (gain: 0.08, SHAP: 0.09), and A35 (gain: 0.05, SHAP: 0.06) dominating the attribution according to both XGBoost gain and SHAP importance. This suggests that irrigation-related information is compressed within the learned embedding space but not explicitly represented by interpretable variables.

For Raw Optical + SAR features, moisture-sensitive spectral regions and Sentinel-1 variables were among the most influential predictors. Shortwave infrared reflectance, vegetation-sensitive bands, and SAR backscatter variables showed high attribution values, consistent with their sensitivity to canopy water content and structural differences between irrigated and non-irrigated fields.

The spectral-index representation emphasized moisture and vegetation indices, particularly NDMI and vegetation-related indicators, demonstrating the importance of water availability and crop condition for irrigation discrimination.

The overall ranking patterns were generally consistent between SHAP attribution and XGBoost feature importance, although minor differences occurred due to their different interpretation mechanisms. SHAP provided a prediction-oriented view by measuring the contribution of each variable to model output, while XGBoost importance reflected the variables most frequently used during tree construction.

*Table 9: XGBoost-based feature importance ranking for irrigation classification across different Earth-observation representations.*

| | AlphaEarth | Raw Optical + SAR | Spectral Indices |
|---|---|---|---|
| Feature Importance | 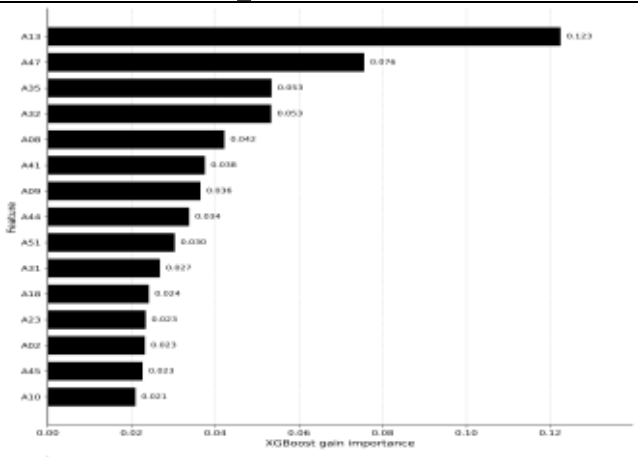 | 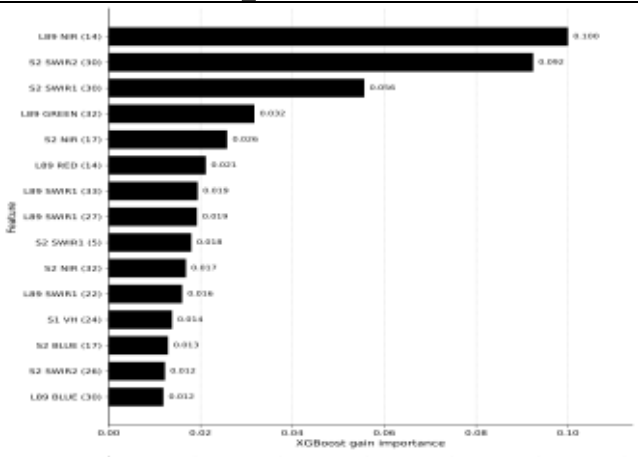 | 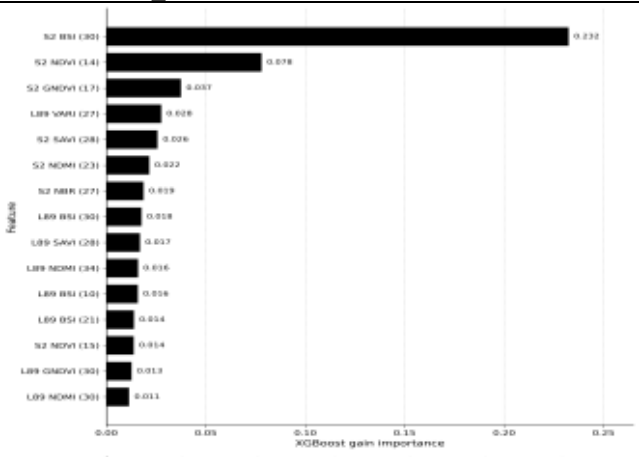 |
| SHAP | 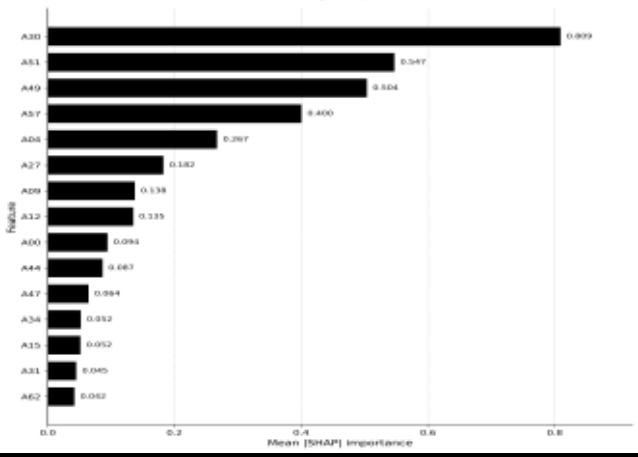 | 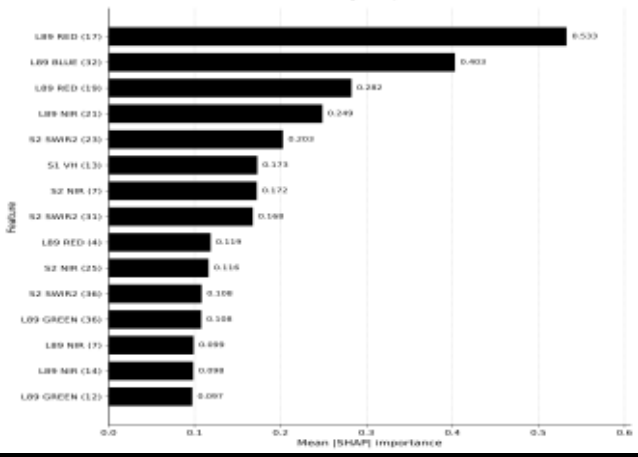 | 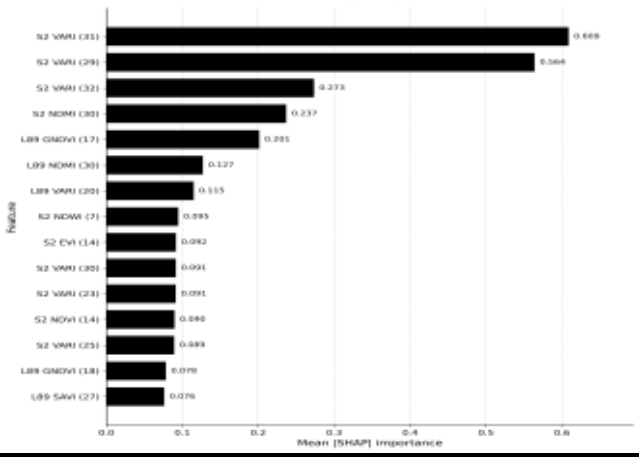 |

The agreement between both approaches increased confidence that the identified variables represented stable predictive drivers rather than artifacts of a single attribution method.

For conventional EO representations, the identified drivers could be directly interpreted through known remote-sensing relationships. For example, vegetation indices represented canopy condition, SWIR bands represented moisture sensitivity, and SAR variables captured structural and dielectric differences. In contrast, the influential AlphaEarth dimensions could only be interpreted as latent predictors, motivating an additional physical probing analysis to investigate whether these dimensions preserve recoverable Earth-observation information.

## 5.5 Physical information recoverability from AlphaEarth embeddings

The results demonstrate substantial recoverability of temporally integrated EO information from the AlphaEarth embeddings. The highest reconstruction performance was obtained for mean SWIR and moisture-sensitive variables. Mean Sentinel-2 SWIR1 achieved an ($R^2$) of 0.93, followed by Landsat 8/9 NDMI ($R^2$ = 0.93), Landsat SWIR1 ($R^2$ = 0.93), Sentinel-2 SWIR2 ($R^2$ = 0.93), and Landsat SWIR2 ($R^2$ = 0.93). Vegetation-sensitive variables were also strongly recoverable, including Landsat NDVI ($R^2$ = 0.91) and Sentinel-2 NDVI ($R^2$ = 0.90). In addition, the embeddings retained substantial information associated with SAR observations: mean Sentinel-1 VH achieved ($R^2$ = 0.92), RVI ($R^2$ = 0.89), the VH/VV ratio ($R^2$=0.88), and VV ($R^2$ = 0.85) (Table 9).

A notable result was the dominance of the linear probe for temporally averaged variables. After excluding two numerically unstable Sentinel-2 EVI and VARI feature families from the aggregate interpretation, Ridge regression provided the best-performing probe for 30 of 32 mean EO variables. Across these mean variables, the median holdout ($R^2$) was 0.89. This indicates that a large proportion of the information describing average spectral, vegetation, moisture, soil/background, and radar conditions is linearly accessible within the AlphaEarth representation rather than requiring a strongly nonlinear decoder.

The Spearman analysis further revealed recurring associations between particular latent dimensions and related EO variables. For example, dimension A49 showed strong associations with vegetation-sensitive indices, including Landsat NDVI mean ($\rho$ =0.73) and Sentinel-2 NDVI mean ($\rho$ =0.72). Dimension A38 was most strongly associated with mean NDMI for both Landsat ($\rho$ =-0.72) and Sentinel-2 ($\rho$ =-0.72), while also showing a strong relationship with Landsat BSI ($\rho$ =0.68). Similarly, A30 was repeatedly associated with SWIR information, whereas A25 was strongly associated with Sentinel-1 VH ($\rho$ =0.71), RVI ($\rho$ =0.70), and VH/VV ($\rho$ =0.69). These recurring cross-variable associations indicate that the latent embedding space is not entirely unstructured with respect to conventional EO observables. However, these relationships should be interpreted as statistical associations and information recoverability rather than as evidence that individual AlphaEarth dimensions possess unique or explicit physical semantics.

A clear contrast emerged between the recoverability of temporally integrated surface conditions and the timing of temporal events. Across the stable feature families, median ($R^2$) values were approximately 0.90 for temporal means, 0.85 for the 10th percentile, 0.81 for the 90th percentile, 0.75 for minima, 0.73 for standard deviation, and 0.626 for temporal amplitude. In contrast, peak timing showed substantially lower recoverability, with a median ($R^2$) of only 0.20. For example, Sentinel-2 NDVI mean was recovered with ($R^2$=0.90), whereas the timing of its maximum response achieved only ($R^2$=0.23). Similarly, Landsat NDVI peak timing reached ($R^2$=0.20), and Sentinel-1 VV peak timing reached only ($R^2$=0.15).

This divergence suggests that AlphaEarth embeddings preserve the integrated spectral and biophysical state of agricultural parcels considerably better than the precise temporal phase of seasonal dynamics. In other words, the representation contains strong information about typical vegetation, moisture, spectral, soil, and radar conditions, but substantially less information about exactly when seasonal extrema occur. This finding is consistent with the known characteristics of AlphaEarth embeddings. The model's self-supervised pretraining reconstructs held-out spectral channels and climate values from the latent representation, prioritizing the encoding of surface reflectance, radar backscatter, and biophysical properties (Wang et al., 2026). However, the annual composite nature of the embeddings limits their capacity to capture intra-seasonal phenological dynamics. Studies have demonstrated that AlphaEarth embeddings can retain 98% of baseline land cover classification accuracy using as few as 2 to 12 of the 64 dimensions, confirming that information is highly compressed but not evenly distributed across temporal characteristics (Ma et al., 2026b). Furthermore, the embedding dimensions themselves lack explicit temporal modeling during pretraining, as the continuous-time STP encoder is trained on annual archives rather than explicit phenological trajectories (Ball et al., 2026b). This behavior provides a possible explanation for the remaining performance advantage of explicit multi-temporal satellite representations in agricultural classification, where crop development and irrigation practices can depend strongly on phenological timing.

*Table 10: Representative physical-probing results showing recoverability of conventional EO variables from the 64-dimensional AlphaEarth representation. The reported embedding dimension corresponds to the strongest absolute Spearman association for each physical variable.*

| **Information domain** | **Physical variable** | **Top AlphaEarth dimension** | **Spearman (ρ)** | **Best probe** | **Holdout ($R^2$)** |
|---|---|---|---|---|---|
| Spectral/moisture-sensitive | S2 SWIR1 mean | A30 | 0.60 | Ridge | **0.93** |
| Moisture | L89 NDMI mean | A38 | -0.72 | Ridge | **0.93** |
| Spectral / moisture-sensitive | L89 SWIR1 mean | A30 | 0.58 | Ridge | **0.93** |
| Spectral / moisture-sensitive | S2 SWIR2 mean | A30 | 0.67 | Ridge | **0.93** |
| SAR structure | S1 VH mean | A25 | 0.71 | Ridge | **0.92** |
| Soil/background | L89 BSI mean | A38 | 0.68 | Ridge | **0.92** |
| Moisture | S2 NDMI mean | A38 | -0.72 | Ridge | **0.91** |
| Vegetation | L89 NDVI mean | A49 | 0.73 | Ridge | **0.91** |
| Vegetation | S2 NDVI mean | A49 | 0.77 | Ridge | **0.20** |
| SAR vegetation/structure | S1 RVI mean | A25 | 0.70 | Ridge | **0.89** |
| SAR structure | S1 VH/VV mean | A25 | 0.69 | Ridge | **0.88** |
| SAR backscatter | S1 VV mean | A49 | 0.63 | Ridge | **0.85** |
| Temporal timing | S2 NDVI peak timing | A41 | 0.31 | Random Forest | **0.23** |
| Temporal timing | L89 NDVI peak timing | A13 | 0.30 | Random Forest | **0.20** |
| Temporal timing | S1 VV peak timing | A04 | 0.22 | Random Forest | **0.15** |

Overall, these results indicate that AlphaEarth embeddings should not be considered physically uninterpretable in the sense of containing no recognizable EO information. Rather, physically meaningful information is strongly recoverable from the latent representation, particularly for temporally averaged spectral, moisture, vegetation, soil, and SAR characteristics. The principal interpretability limitation therefore lies in the absence of explicit one-to-one physical semantics for individual embedding dimensions and in the weaker preservation of precise temporal or phenological timing.

## 6 Practical Implications, Limitations, and Future Research Directions

### 6.1 Practical implications for operational deployment

The findings of this study have several implications for the operational deployment of LULC and irrigation classification systems based on TabPFN and the evaluated feature configurations. In particular, they clarify when the higher computational cost of time-series inputs is justified and when embedding-based representations are preferable from an efficiency perspective.

When maximum classification accuracy is critical, such as in water resources management, agricultural policy formulation, or food security assessments, time-series configurations are recommended despite their higher computational cost. In these settings, the indices time series configuration is especially attractive because it provides a more favorable accuracy-efficiency balance than the surface reflectance time series. In national-scale agricultural monitoring programs that require accurate and consistent land cover and crop-type mapping over large areas, small gains in accuracy (on the order of 2.2-3.0%) can translate into meaningful improvements in the reliability of agricultural statistics, thereby supporting better-informed policy decisions and resource allocation.

In water resource allocation and irrigation management, accurate identification of irrigated areas is essential for optimizing water distribution, improving irrigation efficiency, and ensuring sustainable use. The higher accuracies observed for time-series configurations (e.g., 94.4% for LULC and 87.2% for irrigation) enable more precise estimation of irrigation water demand, which can improve planning and reduce the risk of over-extraction or inefficient allocation. Similarly, food security early warning systems depend on timely and reliable information about crop conditions, irrigation status, and production levels to anticipate and mitigate potential shortages. The improved classification performance of time-series configurations enhances the robustness of early warning indicators, enabling more targeted and timely interventions to prevent food crises.

Finally, climate change adaptation assessments require detailed and accurate baseline information on land cover, irrigation practices, and agricultural systems to evaluate vulnerabilities and design effective adaptation strategies. In this context, the high accuracy of time-series configurations ensures that assessments are grounded in reliable spatial information, supporting the development of robust adaptation measures to address climate-related impacts on agricultural systems.

For applications that require rapid model updates, large-scale processing, near-real-time monitoring, or operation in resource-constrained environments, the embedding configuration with TabPFN provides a practical and efficient solution. With a training time of only 277 seconds and competitive accuracies (90.2% for LULC and 85.0% for irrigation), this setup offers a scalable alternative that maintains strong performance while substantially reducing computational demand. Consequently, this configuration is well suited to use cases such as near-real-time crop monitoring and early warning systems, as well as large-scale continental or global mapping efforts. In such contexts, the substantial computational savings, combined with competitive accuracy, render the embedding configuration an attractive choice when efficiency and scalability are prioritized over marginal gains in classification performance.

### 6.2 Limitations of the Study

While this study provides valuable insights into the performance of different feature configurations for LULC and Irrigation classification, several limitations should be acknowledged.

#### A. Limitations of time-series configurations

Despite the superior classification performance of time-series configurations, they present several significant challenges that limit their operational applicability and motivate the development of embedding-based alternatives.

1) Computational and resource demands

The primary limitation of time-series configurations is their substantial computational cost. The SR-Time Series configuration requires 4,846 seconds of training time, 17.5 times longer than the Embedding configuration (277 seconds). This dramatic increase in computational demand has direct implications for operational deployment: it requires investment in high-performance computing infrastructure, increases energy consumption, and extends model development and retraining cycles. For organizations with limited computational resources, this can be a prohibitive barrier to adoption.

2) Preprocessing complexity

Time-series configurations demand extensive preprocessing of multi-temporal satellite data. The construction of consistent time-series requires harmonization of observations from different sensors, correction of atmospheric effects, gap-filling of missing data due to cloud cover, and alignment of temporal resolutions. This preprocessing pipeline is labor-intensive, requires specialized expertise, and introduces potential sources of error that can propagate through the classification workflow. The complexity of this pipeline increases with the number of time steps and the diversity of input features, making it difficult to scale to large geographic areas or multiple regions.

3) Data availability and consistency

The performance of time-series configurations depends critically on the availability of consistent, high-quality multi-temporal observations. Cloud cover, sensor outages, and data gaps can significantly degrade the quality of time-series features, particularly in regions with frequent cloud cover or limited satellite coverage. This is especially problematic for operational systems that require reliable, near-real-time updates. The dependence on consistent multi-temporal data also limits the transferability of time-series models to regions where such data are not available.

4) Storage requirements

Time-series configurations require storage of multi-temporal satellite data, which can be substantial in volume. For large-scale applications, the storage requirements for time-series data can exceed available capacity, necessitating investment in data storage infrastructure. This adds to the overall cost and complexity of operational systems.

## B. Limitations of the Embedding configuration

While the Embedding configuration offers significant advantages in terms of computational efficiency and operational simplicity, it also presents several limitations that must be acknowledged and addressed in future work.

1) Temporal information loss

The most fundamental limitation of the Embedding configuration is the absence of explicit temporal information. Embeddings are derived from static representations that aggregate multi-temporal observations into a single feature vector. This aggregation inevitably loses the sequential dynamics, phenological patterns, and temporal evolution that are critical for distinguishing between land cover types and irrigation practices. As a result, embeddings cannot capture seasonal variations, growth cycles, or the timing of irrigation events, which are essential discriminative features for agricultural classification.

2) Interpretability and physical meaning

The dimensions of the embedding space are learned through self-supervised pretraining and do not correspond directly to physically interpretable variables. Unlike indices such as NDVI or NDWI, which have clear biophysical meanings, the individual embedding dimensions are latent features with no direct physical interpretation. This lack of interpretability limits the ability to understand why a particular classification decision was made, which can be problematic for applications requiring transparency, accountability, or scientific validation.

3) Contextual dependency

Despite being trained on global data, the performance of embeddings is still dependent on the training data distribution and may not generalize equally well to all regions or conditions. The current study demonstrates that embeddings lag behind time-series configurations by approximately 2.2-3.0% in classification accuracy, suggesting that the learned representations do not fully capture all the discriminative information present in the time-series data. This performance gap highlights the need for improved representation learning that can more effectively encode temporal information.

## 6.3 Future research directions and goals

The limitations of both time-series and embedding configurations suggest several directions for future research that could improve the performance, efficiency, and applicability of these approaches.

The ultimate goal is to create embedding configurations that perform as well as time-series approaches in terms of accuracy, while keeping all the advantages of embeddings (speed, simplicity, and scalability). This would provide a practical, cost-effective solution for large-scale LULC and Irrigation monitoring, making advanced classification tools accessible to a wider range of users and organizations. Adding time-series information to embeddings is the most promising way to achieve this goal, as it addresses the main weakness of current embeddings while building on their strengths.

1) Adding time to embeddings.

Current embeddings are static, they capture information from a single point in time or an average over time, which means they miss important seasonal changes and growth patterns. Future work should develop time-aware embeddings that include information about when things happen, such as when crops grow, when irrigation occurs, or how vegetation changes throughout the year. This could be done by using models that process sequences of observations, like time-series transformers or recurrent neural networks, which can learn patterns over time and encode them into the embedding.

2) Hybrid approaches

Another promising idea is to combine the best of both worlds: use fast, simple embeddings and add a few key time-series features, like seasonal indices or phenological metrics. This hybrid approach would give us the richness of time-series data without the high computational cost.

3) Making embeddings adaptable

Currently, embeddings are fixed and cannot be updated with new information. Future work should develop adaptive embeddings that can be fine-tuned or updated as new data become available. This would make embedding-based systems more flexible and better able to handle changes in conditions or new regions.

# 7 Conclusion

To evaluate the applicability of AEF embeddings for operational agricultural monitoring, we assessed their performance in terms of classification accuracy, training efficiency, and feature engineering requirements. This paper systematically compares 64-dimensional AEF embeddings against four traditional feature sets, single-date surface reflectance, single-date spectral indices, surface reflectance time series, and spectral index time series. Using XGBoost, CatBoost, and TabPFN with LOOV and 5CV validation protocols on 9,73 labelled parcels from Moroccan irrigated perimeters from 2023 to 2025. Multi-temporal time series consistently outperformed all other feature modalities, achieving the highest classification accuracy (93.5% for LULC and 87.0% for irrigation with TabPFN), underscoring the critical role of phenological information in agricultural discrimination. AEF embeddings demonstrated near-time-series performance (90.6% LULC, 85.6% irrigation) without manual feature engineering, with training time reduced from up to 4,846 seconds to 277 seconds. Spatial LULC mapping using TabPFN achieved comparable overall accuracies (0.85-0.86) across the three feature configurations, but revealed substantial differences in mapped tree cover, ranging from 14.57% for surface reflectance to 29.79% for AFE. These findings highlight the influence of feature representation on LULC spatial estimates despite similar classification performance. Explainability analysis revealed that AEF embeddings concentrate predictive information in limited latent dimensions, with specific dimensions strongly encoding SWIR (A30, $\rho = 0.67$), moisture (A38, $\rho = -0.72$), vegetation (A49, $\rho = 0.73$), and SAR structure (A25, $\rho = 0.71$), but capture phenological timing less directly ($R^2 < 0.23$). These findings confirm that AEF embeddings offer a viable, efficient, and physically interpretable alternative to conventional feature engineering, particularly suitable for resource-constrained and large-scale agricultural monitoring applications, while highlighting the need for improved temporal encoding in future foundation model designs.

**Acknowledgement.**

**Conflicts of Interest.** The authors declare that they have no known competing financial interests or personal relationships that could have appeared to influence the work reported in this paper.

**Funding.** No funding was received to assist with the preparation of this manuscript.

**Use of Generative AI tools:** During the preparation of this manuscript, we utilized ChatGPT to enhance the language clarity and readability of the text. All content generated by the AI tool was rigorously reviewed and edited to ensure scientific accuracy and adherence to the original research intent.

## References

Alvarez, C.I., Govind, A., 2025. Assessing climate and land use changes in Morocco (2001–2023): from a geospatial and farmers' perspective. Theoretical and Applied Climatology 156. https://doi.org/10.1007/s00704-025-05656-z

Amani, M., Ghorbanian, A., Ahmadi, S.A., Kakooei, M., Moghimi, A., Mirmazloumi, S.M., Moghaddam, S.H.A., Mahdavi, S., Ghahremanloo, M., Parsian, S., Wu, Q., Brisco, B., 2020. Google Earth Engine Cloud Computing Platform for Remote Sensing Big Data Applications: A Comprehensive Review. IEEE Journal of Selected Topics in Applied Earth Observations and Remote Sensing 13, 5326–5350. https://doi.org/10.1109/JSTARS.2020.3021052

Arai, K., Nakaoka, Y., Okumura, H., 2025. Method for Landslide Area Detection with RVI Data Which Indicates Base Soil Areas Changed from Vegetated Areas. Remote Sensing 17. https://doi.org/10.3390/rs17040628

Ball, J.G., Wicklein, J.A., Feng, Z., Knezevic, J., Jaffer, S., Madhavapeddy, A., Atzberger, C., Dalponte, M., Coomes, D., 2026a. Geospatial foundation models enable data-efficient tree species mapping in temperate mountain forests. bioRxiv 14, 2026.02.23.707022. https://doi.org/10.1016/j.srs.2026.100466

Ball, J.G., Wicklein, J.A., Feng, Z., Knezevic, J., Jaffer, S., Madhavapeddy, A., Atzberger, C., Dalponte, M., Coomes, D., 2026b. Geospatial foundation models enable data-efficient tree species mapping in temperate mountain forests. bioRxiv 14, 2026.02.23.707022. https://doi.org/10.1016/j.srs.2026.100466

Bihui Lu, Kun Yu, Qiu, L., Huayong Li, Wang, H., Zhang, W., Xiaohong Liu, Shan, J., Li, N., 2026. Predicting county-level winter wheat yield in Eastern China using multi-source spatiotemporal data: An explainable machine learning approach. Agricultural and Forest Meteorology 12, 25–41.

Bousbih, S., Zribi, M., Hajj, M. El, Baghdadi, N., Lili-Chabaane, Z., Gao, Q., Fanise, P., 2018. Soil moisture and irrigation mapping in a semi-arid region, based on the synergetic use of Sentinel-1 and Sentinel-2 data. Remote Sensing 10, 1–22. https://doi.org/10.3390/rs10121953

Brown, C.F., Kazmierski, M.R., Pasquarella, V.J., Rucklidge, W.J., 2025. AlphaEarth Foundations : An embedding field model for accurate and efficient global mapping from sparse label data.

Chen, Q., Vaudour, E., Richer-de-Forges, A.C., Arrouays, D., 2025. Spectral indices in remote sensing of soil: definition, popularity, and issues. A critical overview. Remote Sensing of Environment 329, 114918. https://doi.org/10.1016/j.rse.2025.114918

Chen, T., Guestrin, C., 2016. XGBoost : A Scalable Tree Boosting System, in: In Proceedings of the 22nd Acm Sigkdd International Conference on Knowledge Discovery and Data Mining. pp. 785–794.

DeepMind, 2025. Supplemental Training Coordinates. https://doi.org/https://doi.org/10.5281/zenodo.16585910

Dijk, M. Van, Geurtsen, S., 2023. Mapping Irrigated Areas in China Using a Synergy Approach. Water 1–16.

El Hachimi, C., Belaqziz, S., Khabba, S., Chehbouni, A., 2022. Data Science Toolkit: An all-in-one python library to help researchers and practitioners in implementing data science-related algorithms with less effort[Formula presented]. Software Impacts 12, 100240. https://doi.org/10.1016/j.simpa.2022.100240

EL Hachimi, C., Weng, Q., Baiti, J.R., Khabba, S., El Moussaoui, E.H., Belaqziz, S., AIT Hssaine, B., Chehbouni, A., Kong, H., 2026. MAPPING CROP AND IRRIGATION TYPES IN MOROCCO USING A SATELLITE EMBEDDING FOUNDATION MODEL, in: IEEE Mediterranean and Middle-East Geoscience and Remote Sensing Symposium (M2GARSS). IEEE. pp. 255–259.

El Moussaoui, E.H., Moumni, A., Khabba, S., Amazirh, Abdelhakim Er-Raki, S., Abdelghani, Chehbouni Lahrouni, A., 2025. A comparative methodological approach for argan forest classification using Landsat imagery. Environmental Monitoring and Assessment 197:210. https://doi.org/10.1007/s10661-025-13649-8

El Moussaoui, E.H., Moumni, A., Lahrouni, A., 2024. Assessing the influence of different Synthetic Aperture Radar parameters and Digital Elevation Model layers combined with optical data on the identification of argan forest in Essaouira region , Morocco. iForest - Biogeosciences and Forestry 17, 100–108. https://doi.org/10.3832/ifor4183-016

El-bouhali, A., Cotonnec, A., Lebaut, S., Amyay, M., 2023. Remote sensing monitoring of irrigated areas from 1972 to 2018 in the Guigou Plain , Middle Atlas , Morocco. Journal of water and land development 56, 249–261. https://doi.org/10.24425/jwld.2023.143766

Elwan, E., Page, M. Le, Jarlan, L., Baghdadi, N., Brocca, L., Modanesi, S., Dari, J., Segui, P.Q., Zribi, M., 2022. Irrigation Mapping on Two Contrasted Climatic Contexts Using Sentinel-1 and Sentinel-2 Data. Water (Switzerland) 14, 1–13. https://doi.org/10.3390/w14050804

FAO, 2020. Morocco water sector brief.

Feng, Z., Atzberger, C., Jaffer, S., Knezevic, J., Sormunen, S., Young, R., Lisaius, M.C., Immitzer, M., Jackson, T., Ball, J., Coomes, D.A., Madhavapeddy, A., Blake, A., Keshav,

S., 2026. TESSERA: Temporal Embeddings of Surface Spectra for Earth Representation and Analysis.

Fernández-Buces, N., Siebe, C., Cram, S., Palacio, J.L., 2006. Mapping soil salinity using a combined spectral response index for bare soil and vegetation: A case study in the former lake Texcoco, Mexico. Journal of Arid Environments 65, 644–667. https://doi.org/10.1016/j.jaridenv.2005.08.005

Gao, B.-C., 1996. NDWI A Normalized Difference Water Index for Remote Sensing of Vegetation Liquid Water From Space. Remote Sensing of Environment 58, 257–266. https://doi.org/https://doi.org/10.1016/S0034-4257(96)00067-3

Gao, Q., Zribi, M., Escorihuela, M.J., 2017. Synergetic Use of Sentinel-1 and Sentinel-2 Data for Soil Moisture Mapping at 100 m Resolution. sensors. https://doi.org/10.3390/s17091966

Gao, Q., Zribi, M., Escorihuela, M.J., Baghdadi, N., Segui, P.Q., 2018. Irrigation mapping using Sentinel-1 time series at field scale. Remote Sensing 10, 1–18. https://doi.org/10.3390/rs10091495

García-Ontiyuelo, M., Acuña-Alonso, C., Peco-Costas, C., Álvarez, X., 2026. Impact of atmospheric correction on machine learning-based land cover classification: insights from the Sil Canyon (NW Spain). Advances in Space Research 77, 8559–8579. https://doi.org/10.1016/j.asr.2026.02.055

Gitelson, A.A., Mark N, M., 1998. Remote_sensing_of_chlorophyll_concentrat.pdf. Remote Sensing of Environment 22, 689–692.

Gitelson, A.A., Stark, R., Rundquist, D., Gitelson, A.A., Kaufman, Y.J., Stark, R., Rundquist, D., 2002. DigitalCommons @ University of Nebraska - Lincoln Novel Algorithms for Remote Estimation of Vegetation Fraction. Natural Resources.

Griffiths, M.J.D., McAuley, D.F., Perkins, G.D., Barrett, N., Blackwood, B., Boyle, A., Chee, N., Connolly, B., Dark, P., Finney, S., Salam, A., Silversides, J., Tarmey, N., Wise, M.P., Baudouin, S. V., 2019. Guidelines on the management of acute respiratory distress syndrome. BMJ Open Respiratory Research 6. https://doi.org/10.1136/bmjresp-2019-000420

Halder, K., Ewert, F., Ghosh, A., Muduchuru, K., Sweet, L. belle, Elshawi, R., Timko, J., Zheng, W., Alsafadi, K., Zhao, G., Maerker, M., Singh, M., Guoging, L., Gaiser, T., Behrend, D., Shi, Y., Han, L., Ryo, M., Srivastava, A.K., 2026. High-resolution maize yield mapping across Africa using earth observation and machine learning, deep learning, and foundation model. Science of Remote Sensing 13, 100344. https://doi.org/10.1016/j.srs.2025.100344

Hollmann, N., Eggensperger, K., Hutter, F., 2023a. TABPFN : A TRANSFORMER THAT SOLVES SMALL TABULAR CLASSIFICATION PROBLEMS IN A SECOND, in: Arxiv.

Hollmann, N., Müller, S., Hutter, F., 2023b. Large Language Models for Automated Data Science: Introducing CAAFE for Context-Aware Automated Feature Engineering.

Advances in Neural Information Processing Systems 36, 1–23. https://doi.org/10.52202/075280-1938

Hoque, O.B., Mandal, N.C., Adiga, A., Swarup, S., Nouwakpo, S.K., Wilson, A., Marathe, M., 2025. Knowledge-Informed Deep Learning for Irrigation Type Mapping from Remote Sensing. IJCAI International Joint Conference on Artificial Intelligence 9692–9700. https://doi.org/10.24963/ijcai.2025/1077

Huete, A., Didan, K., Miura, T., Rodriguez, E.P., Gao, X., Ferreira, L.G., 2002. Overview of the radiometric and biophysical performance of the MODIS vegetation indices. Remote Sensing of Environment 83, 195–213. https://doi.org/10.1016/S0034-4257(02)00096-2

Huete, A.R., 1988. A soil-adjusted vegetation index (SAVI). Remote Sensing of Environment 25, 295–309. https://doi.org/10.1016/0034-4257(88)90106-X

Iounousse, J., Er-raki, S., Motassadeq, A. El, Chehouani, H., 2015. Using an unsupervised approach of Probabilistic Neural Network ( PNN ) for land use classification from multitemporal satellite images 1–13.

Jakubik, J., Yang, F., Blumenstiel, B., Scheurer, E., Sedona, R., Maurogiovanni, S., Bosmans, J., Dionelis, N., Marsocci, V., Kopp, N., Ramachandran, R., Fraccaro, P., Brunschwiler, T., Cavallaro, G., Bernabe-Moreno, J., Longépé, N., 2026. TerraMind: Large-Scale Generative Multimodality for Earth Observation 7383–7394. https://doi.org/10.1109/iccv51701.2025.00693

Jin, S., Sader, S.A., 2005. Comparison of time series tasseled cap wetness and the normalized difference moisture index in detecting forest disturbances. Remote Sensing of Environment 94, 364–372. https://doi.org/10.1016/j.rse.2004.10.012

Li, W., Xiao, C., Liang, X., Yang, W., Zhang, J., Dai, R., La, Y., Kang, L., Zhao, D., 2025. Precision Identification of Irrigated Areas in Semi-Arid Regions Using Optical-Radar Time-Series Features and Ensemble Machine Learning. Hydrology 12, 1–23. https://doi.org/10.3390/hydrology12080214

Lu, S., Guo, J., Zimmer-Dauphinee, J.R., Nieusma, J.M., Wang, X., VanValkenburgh, P., Wernke, S.A., Huo, Y., 2025. Vision Foundation Models in Remote Sensing: A survey. IEEE Geoscience and Remote Sensing Magazine 13, 190–215. https://doi.org/10.1109/MGRS.2025.3541952

Lundberg, S.M., Lee, S.-I., 2017. A Unified Approach to Interpreting Model Predictions, in: Advances in Neural Information Processing Systems.

Ma, Y., Shen, Y., Swatantran, A., Lobell, D.B., 2026a. Harvesting AlphaEarth: Benchmarking the geospatial foundation model for agricultural downstream tasks. International Journal of Applied Earth Observation and Geoinformation 149, 105258. https://doi.org/10.1016/j.jag.2026.105258

Ma, Y., Shen, Y., Swatantran, A., Lobell, D.B., 2026b. Harvesting AlphaEarth: Benchmarking the geospatial foundation model for agricultural downstream tasks. International Journal of Applied Earth Observation and Geoinformation 149, 105258. https://doi.org/10.1016/j.jag.2026.105258

Massari, C., Modanesi, S., Dari, J., Gruber, A., De Lannoy, G.J.M., Girotto, M., Quintana-Seguí, P., Le Page, M., Jarlan, L., Zribi, M., Ouaadi, N., Vreugdenhil, M., Zappa, L., Dorigo, W., Wagner, W., Brombacher, J., Pelgrum, H., Jaquot, P., Freeman, V., Volden, E., Prieto, D.F., Tarpanelli, A., Barbetta, S., Brocca, L., 2021. A review of irrigation information retrievals from space and their utility for users. Remote Sensing 13, 1–26. https://doi.org/10.3390/rs13204112

Modanesi, S., Busschaert, L., De Lannoy, G., De Santis, D., Natali, M., Dari, J., Quintana-Seguí, P., Castelli, M., Grasso, F.M., Massari, C., 2025. Accounting for scaling effects on irrigation optimization within a land surface model using satellite observations. Journal of Hydrometeorology. https://doi.org/10.1175/jhm-d-25-0057.1

Morell-Monzó, S., Sebastiá-Frasquet, M.T., Estornell, J., Moltó, E., 2023. Detecting abandoned citrus crops using Sentinel-2 time series. A case study in the Comunitat Valenciana region (Spain). ISPRS Journal of Photogrammetry and Remote Sensing 201, 54–66. https://doi.org/10.1016/j.isprsjprs.2023.05.003

Moumni, A., Lahrouni, A., 2021. Machine Learning-Based Classification for Crop-Type Mapping Using the Fusion of High-Resolution Satellite Imagery in a Semiarid Area. Scientific 2021. https://doi.org/10.1155/2021/8810279

Nouwakpo, S.K., Bjorneberg, D., McGwire, K., Hoque, O., 2024. Mapping Irrigation Methods in the Northwestern US Using Deep Learning Classification. Water Resources Research 60. https://doi.org/10.1029/2023WR036155

Olofsson, P., Foody, G.M., Herold, M., Stehman, S. V, Woodcock, C.E., Wulder, M.A., 2014. Good practices for estimating area and assessing accuracy of land change. Remote Sensing of Environment 148, 42–57. https://doi.org/10.1016/j.rse.2014.02.015

Ouaadi, N., Jarlan, L., Ezzahar, J., Zribi, M., Khabba, S., Bouras, E., Bousbih, S., Frison, P.L., 2020. Monitoring of wheat crops using the backscattering coefficient and the interferometric coherence derived from Sentinel-1 in semi-arid areas. Remote Sensing of Environment 251, 112050. https://doi.org/10.1016/j.rse.2020.112050

Ouassanouan, Y., Elfarkh, J., Grich, S., Liblab, A., Chehbouni, A., 2026. Crop and irrigation types ground-truth dataset for Moroccan agricultural regions. Data, Scientific. https://doi.org/https://doi.org/10.1038/ s41597-026-06993-y

Ouassanouan, Y., Fakir, Y., Simonneaux, V., Hakim, M., Bouimouass, H., Najar, I., Benrhanem, M., Sguir, F., Ouassanouan, Y., Fakir, Y., Simonneaux, V., Kharrou, M.H., Bouimouass, H., 2023. Multi-decadal analysis of water resources and agricultural change in a Mediterranean semiarid irrigated piedmont under water scarcity and human interaction To cite this version : HAL Id : ird-04351656. Science of the Total Environment,.

Page, M. Le, Jarlan, L., Hajj, M.M. El, Zribi, M., Baghdadi, N., Boone, A., 2020. Potential for the Detection of Irrigation Events on Maize Plots Using Sentinel-1 Soil Moisture Products. remote sensing Article 1–22.

Prokhorenkova, L., Gusev, G., Vorobev, A., Dorogush, A.V., Gulin, A., 2018. CatBoost : unbiased boosting with categorical features. Advances in neural information processing systems 1–11.

Qi, J., Chehbouni, A., Huete, A.R., Kerr, Y.H., Sorooshian, S., 1994. A modified soil adjusted vegetation index. Remote Sensing of Environment 48, 119–126. https://doi.org/10.1016/0034-4257(94)90134-1

Rahman, M., 2026. Physically interpretable AlphaEarth foundation model embeddings enable LLM-based land surface intelligence. Remote Sensing Applications: Society and Environment 42, 102045. https://doi.org/10.1016/j.rsase.2026.102045

ROPUR, 2025. Water scarcity in Morocco: 2024–2028 challenges and solutions. [WWW Document]. Toray Membrane Europe.

Schmidinger, J., Barkov, V., Vogel, S., Atzmueller, M., Heuvelink, G.B.M., 2026. Kriging prior regression: A case for kriging-based spatial features with TabPFN in soil mapping. Computers and Electronics in Agriculture 243. https://doi.org/10.1016/j.compag.2025.111352

Shamal, S.A.M., Weatherhead, K., 2014. Assessing spectral similarities between rainfed and irrigated croplands in a humid environment for irrigated land mapping. Outlook on Agriculture 43, 109–114. https://doi.org/10.5367/oa.2014.0168

Stehman, S. V., Foody, G.M., 2019. Key Issues in Rigorous Accuracy Assessment of Land Cover Products. Remote Sensing of Environment, 231, 111199.

Sun, Y., Kosmas, P., 2023. Meta-forests: Domain generalization on random forests with meta-learning. Proceedings of Machine Learning Research 222, 1292–1307.

Tabasum, M., Tasnim, T., Islam, Md.E., Habib, A.Z.S. Bin, 2025. Tabular Deep Learning vs Classical Machine Learning for Urban Land Cover Classification, in: In 5th Muslims in ML Workshop Co-Locate.

Tucker, C.J., 1979. Red and photographic infrared linear combinations for monitoring vegetation. Remote Sensing of Environment 8, 127–150. https://doi.org/10.1016/0034-4257(79)90013-0

Veloso, A., Mermoz, S., Bouvet, A., Le Toan, T., Planells, M., Dejoux, J.F., Ceschia, E., 2017. Understanding the temporal behavior of crops using Sentinel-1 and Sentinel-2-like data for agricultural applications. Remote Sensing of Environment 199, 415–426. https://doi.org/10.1016/j.rse.2017.07.015

Vermote, E., Justice, C., Claverie, M., Franch, B., 2016. Preliminary analysis of the performance of the Landsat 8/OLI land surface reflectance product. Remote Sens Environ 185, 46–56. https://doi.org/10.1016/j.rse.2016.04.008.Preliminary

Vinge, R., Marszalek, M.L., Schneider, J., Albrecht, C.M., 2025. Earth Observation embeddings at the test: A novel benchmark to evaluate (neural) compression for satellite imagery, in: EGU. https://doi.org/https://doi.org/10.5194/egusphere-egu25-16756, 2025

Wang, Y., Zhang, X., Gao, G., Liu, L., 2026. Evaluating the Efficacy of AlphaEarth Embeddings for Tree Species Mapping in Complex Terrain. IEEE Journal of Selected Topics in Applied Earth Observations and Remote Sensing 19, 19719–19733. https://doi.org/10.1109/JSTARS.2026.3699640

Xiao, A., Xuan, W., Wang, J., Huang, J., Tao, D., Lu, S., Yokoya, N., 2025. Foundation Models for Remote Sensing and Earth Observation: A survey. IEEE Geoscience and Remote Sensing Magazine 13, 297–324. https://doi.org/10.1109/MGRS.2025.3576766

Yang, L., Gao, Y., Zhao, X., Liang, N., Ma, R., Xi, S., Zhang, X., Wang, R., 2026. Evaluating the Performance of AlphaEarth Foundation Embeddings for Irrigated Cropland Mapping Across Regions and Years. Remote Sensing 18, 1–24. https://doi.org/10.3390/rs18071065

Yousaf, W., Ahmad, S.R., Shahzad, N., Ramzan, A., Javaid, A., 2025. An Object-Based Crop Classification Using Optimum Remotely Sensed Phenological and Multi-Spectral Data in Pakistan. Remote Sensing in Earth Systems Sciences 8, 945–964. https://doi.org/10.1007/s41976-025-00229-0

Zheng, C., Jia, L., Menenti, M., Hu, G., Lu, J., Chen, Q., Jiang, M., Mancini, M., Corbari, C., 2024. Integrating hydrologic modeling and satellite remote sensing to assess the performance of sprinkler irrigation. Geo-Spatial Information Science 27, 934–952. https://doi.org/10.1080/10095020.2024.2344615

Zheng, X., Jia, J., Dong, S., Wang, Yawei, Lu, R., Chen, Y., Wang, Yueming, 2025. Training and inference Time Efficiency Assessment Framework for machine learning algorithms: A case study for hyperspectral image classification. International Journal of Applied Earth Observation and Geoinformation 141, 104591. https://doi.org/10.1016/j.jag.2025.104591

Zvonkov, I., Tseng, G., Becker-Reshef, I., Kerner, H., 2025. Cropland Mapping using Geospatial Embeddings. arXiv preprint arXiv:2511, 1–8.